\documentclass[runningheads]{llncs}
\usepackage{graphicx}
\usepackage{amsmath,amssymb} 
\usepackage{xcolor}
\usepackage[width=122mm,left=12mm,paperwidth=146mm,height=193mm,top=12mm,paperheight=217mm]{geometry}

\usepackage{subfigure}
\usepackage{pbox}
\usepackage{wrapfig}

\usepackage{booktabs}
\usepackage{array}
\usepackage{xspace}

\usepackage{hyperref}
\hypersetup{
  linkcolor  = black,
  citecolor = black,
  urlcolor   = green!20!black,
  colorlinks = true,
}

\newcommand{\mycaption}[2]{\caption{\small \textbf{#1.}~#2}}
\newcommand{\bff}{\mathbf{f}}
\newcommand{\bz}{\mathbf{z}}
\newcommand{\bw}{\mathbf{w}}

\newcommand{\deconv}{DeconvNet}
\newcommand{\deconvcrf}{DeconvNet-CRF}

\newcommand{\deeplablargefov}{DeepLab}
\newcommand{\deeplablargefovcrf}{DeepLab-CRF}
\newcommand{\deeplabmsclargefovcrf}{DeepLab-MSc-CRF}

\newcommand{\deeplab}{DeepLab}

\DeclareRobustCommand\onedot{\futurelet\@let@token\@onedot}
\def\onedot{\ifx\let@token.\else.\null\fi\xspace}
\def\eg{\emph{e.g}\onedot}

\newcommand{\bi}[2]{BI$_{#1}(#2)$}

\newcommand{\fc}[1]{FC$_{#1}$}

\begin{document}
\pagestyle{headings}
\mainmatter
\def\ECCV16SubNumber{187}  

\title{Superpixel Convolutional Networks using Bilateral Inceptions} 

\titlerunning{Superpixel Convolutional Networks using Bilateral Inceptions}

\authorrunning{Gadde et al.}

\author{Raghudeep Gadde$^{1\star}$, Varun Jampani$^{2\star}$, Martin Kiefel$^{2,3}$, Daniel Kappler$^{2}$, and Peter V. Gehler$^{2,3}$}
\institute{$^{1}$Universit\'e Paris-Est, LIGM (UMR 8049), CNRS, ENPC, ESIEE, UPEM, France \\
$^{2}$Max Planck Institute for Intelligent Systems, T{\"u}bingen, Germany \\
$^{3}$Bernstein Center for Computational Neuroscience, T{\"u}bingen, Germany}

\footnotetext[1]{The first two authors contribute equally to this work.}

\maketitle

\begin{abstract}
  In this paper we propose a CNN architecture for semantic image segmentation. We
  introduce a new ``bilateral inception'' module that can be inserted in
  existing CNN architectures and performs bilateral
  filtering, at multiple feature-scales, between superpixels in an image.
  The feature spaces for bilateral
  filtering and other parameters of the module are learned end-to-end using
  standard backpropagation techniques.
  The bilateral inception module addresses two issues that
  arise with general CNN segmentation architectures. First, this module propagates
  information between (super) pixels while respecting image edges, thus
  using the structured information of the problem for improved results.
  Second, the layer recovers a full resolution segmentation result from the
  lower resolution solution of a CNN.
  In the experiments, we modify several existing CNN architectures by inserting
  our inception module between the last CNN ($1\times1$ convolution) layers. Empirical results
  on three different datasets show reliable improvements not only in comparison
  to the baseline networks, but also in comparison to several dense-pixel prediction
  techniques such as CRFs, while being competitive in time.
\end{abstract}

\section{Introduction}

In this paper we propose a CNN architecture for semantic image segmentation.
Given an image $\mathcal{I}=(x_1,\ldots,x_N$) with $N$ pixels $x_i$ the task of semantic segmentation is to infer a labeling $Y=(y_1,\ldots,y_N)$ with a label $y_i\in\mathcal{Y}$ for every pixel.
This problem can be naturally formulated as a structured prediction problem $g:\mathcal{I}\rightarrow Y$.
Empirical performance is measured by comparing $Y$ to a human labeled $Y^*$ via a
loss function $\Delta(Y,Y^*)$, \eg, with the Intersection over Union (IoU) or pixel-wise
Hamming Loss.

A direct way to approach this problem would be to ignore the structure of the
output variable $Y$ and train a classifier that predicts the class membership of the center
pixel of a given image patch. This procedure reduces the problem to a standard multi-class
classification problem and allows the use of standard learning algorithms. The resulting
classifier is then evaluated at every possible patch in a sliding window fashion (or using
coarse-to-fine strategies) to yield a full segmentation of the image.
With high capacity models and large amounts of training data this approach would
be sufficient, given that the loss decomposes over the pixels.
Such a per-pixel approach ignores the relationship between the variables $(y_1,\ldots,y_N)$, which are not i.i.d.~since there is an underlying common image. Therefore, besides learning
discriminative per-pixel classifiers, most segmentation approaches further encode the output
relationship of $Y$. A dominating approach is to use Conditional Random Fields
(CRF)~\cite{lafferty2001crf}, which allows an elegant and principled way
to combine single pixel predictions and shared structure through unary, pairwise and
higher order factors.

\begin{figure}[t]
\begin{center}
\centerline{\includegraphics[width=0.8\textwidth]{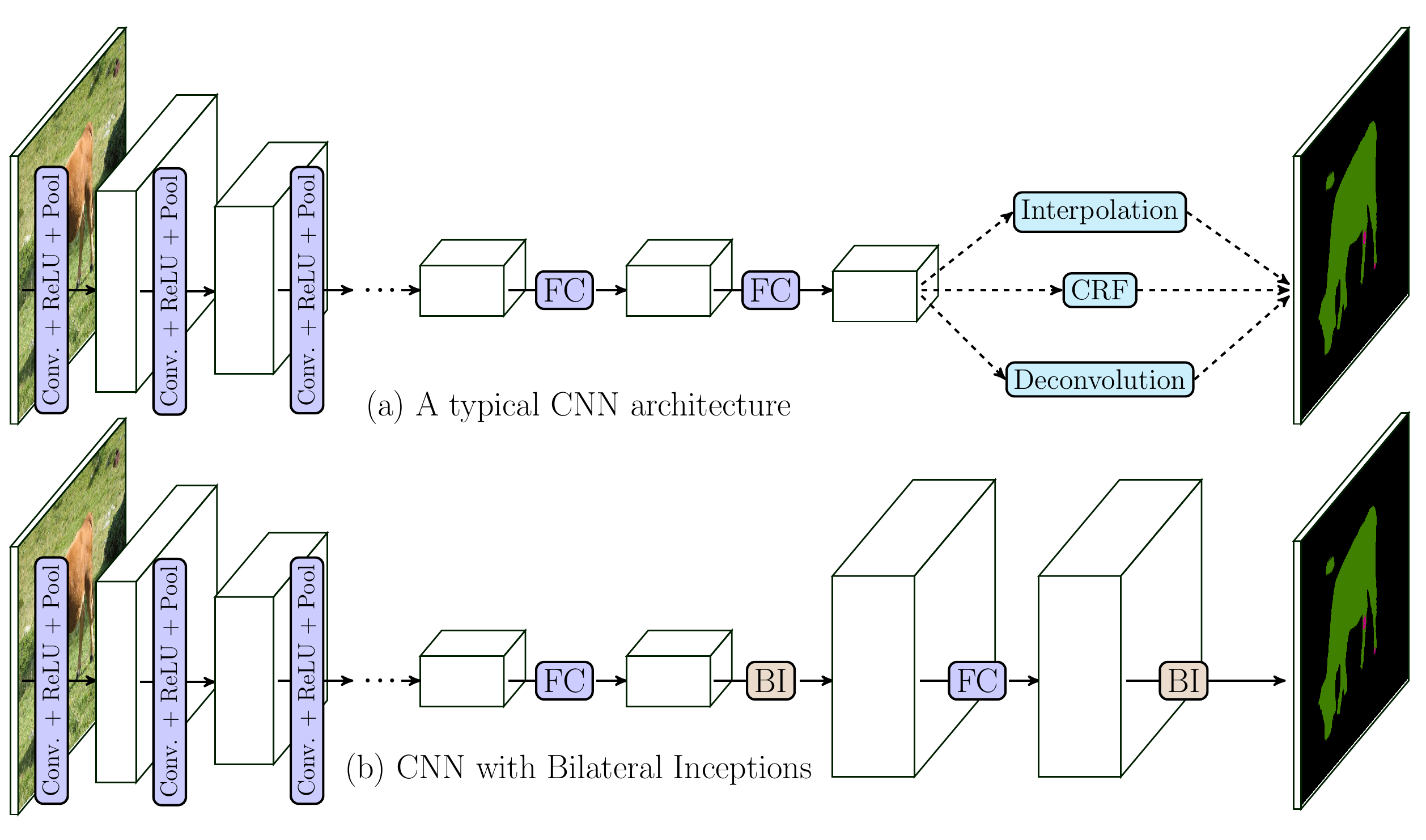}}
  \mycaption{Illustration of CNN layout} {We insert the \emph{Bilateral Inception (BI)} modules
  between the \emph{FC} ($1\times1$ convolution) layers found in
  most networks thus removing the necessity
  of further up-scaling algorithms. Bilateral Inception modules also propagate information
  between distant pixels based on their spatial and color similarity and work
  better than other label propagation approaches.}\label{fig:illustration}
\end{center}
\end{figure}

What relates the outputs $(y_1,\ldots,y_N$)? The common hypothesis that
we use in this paper could be summarized as:
\emph{Pixels that are spatially and photometrically similar are more likely to have the same label.}
Particularly if two pixels $x_i,x_j$ are close in the image and have similar
$RGB$ values, then their corresponding labels $y_i,y_j$ will most likely be the same.
The most prominent example of spatial similarity encoded in a CRF is the Potts model (Ising model for
the binary case).
The work of~\cite{krahenbuhl2012efficient} described a densely connected pairwise CRF (DenseCRF) that includes
pairwise factors encoding both spatial \emph{and} photometric similarity.
The DenseCRF has been used in many recent works on image segmentation
which find also empirically improved results over pure pixel-wise CNN
classifiers~\cite{chen2014semantic,bell2015minc,zheng2015conditional,chen2015semantic}.

In this paper, we implement the above-mentioned hypothesis of photometrically similar
and near-by pixels share common labels, by designing a new
``Bilateral Inception'' (BI) module that can be inserted before/after the last $1\times1$ convolution
layers (which we refer to as `FC' layers - `Fully-Connected' in the original image classification network) of the standard segmentation CNN architectures. The bilateral inception module does edge-aware information propagation across different spatial CNN units of the previous FC layer. Instead of using the spatial grid-layout that is common in CNNs,
we incorporate the superpixel-layout for information propagation. The information
propagation is performed using standard bilateral filters with Gaussian kernels, at
different feature scales. This construction is inspired by~\cite{szegedy2014googlenet,lin2014network}. Feature spaces and other parameters of
the modules can be learned end-to-end using standard backpropagation techniques.
The application of superpixels reduces the number of necessary computations and implements a long-range edge-aware inference between different superpixels. Moreover, since superpixels
provides an output at the full image resolution it removes the need
for any additional post-processing step.

We introduce BI modules in the CNN segmentation models
of~\cite{chen2014semantic,zheng2015conditional,bell2015minc}.
See Fig.~\ref{fig:illustration} for an illustration. This
achieves better segmentation results on all three datasets
we experimented with than the proposed
interpolation/inference techniques of DenseCRF~\cite{bell2015minc,chen2014semantic}
while being faster. Moreover, the results compare favorably against some recently
proposed dense pixel prediction techniques.
As illustrated in Fig.~\ref{fig:illustration}, the BI modules
provides an alternative approach to commonly used up-sampling and CRF
techniques.


\section{Related Work}\label{sec:related}

The literature on semantic segmentation is large and therefore we
will limit our discussion to those works that perform segmentation with
CNNs and discuss the different ways to encode the output structure.

A natural combination of CNNs and CRFs is to use the CNN as unary potential
and combine it with a CRF that also includes pairwise or higher order factors.
For instance~\cite{chen2014semantic,bell2015minc}
observed large improvements in pixel accuracy when combining a DenseCRF~\cite{krahenbuhl2012efficient}
with a CNN. The mean-field steps of the DenseCRF can
be learned and back-propagated as noted by~\cite{domke2013learning} and implemented
by~\cite{zheng2015conditional,arxivpaper,li2014mean,schwing2015fully} for semantic segmentation and~\cite{kiefel2014human} for human pose estimation.
The works of~\cite{chen2014learning,lin2015efficient,liu2015semantic}
use CNNs also in pairwise and higher order factors for more expressiveness.
The recent work of~\cite{chen2015semantic} replaced the costly
DenseCRF with a faster domain transform performing smoothing filtering while
predicting the image edge maps at the same time.
Our work was inspired by DenseCRF approaches but with the aim to replace the
expensive mean-field inference. Instead of propagating information across unaries
obtained by a CNN, we aim to do the edge-aware information propagation across
\textit{intermediate} representations of the CNN. Experiments on different datasets indicate that
the proposed approach generally gives better results in comparison to DenseCRF
while being faster.

A second group of works aims to inject the structural knowledge in intermediate
CNN representations by using structural layers among CNN internal layers.
The deconvolution layers model from~\cite{zeiler2010deconvolutional}
are being widely used for local propagation of information.
They are computationally efficient and are used in segmentation networks, for \eg~\cite{long2014fully}.
They are however limited to small receptive fields. Another
architecture proposed in~\cite{he2014spatial} uses spatial pyramid pooling layers to max-pool
over different spatial scales. The work of~\cite{ionescu2015matrix} proposed specialized
structural layers such as normalized-cut layers with matrix back-propagation techniques.
All these works have either fixed local receptive fields and/or have their complexity increasing exponentially with longer range pixel connections.
Our technique allows for modeling long range (super-) pixel dependencies without compromising the computational efficiency.
A very recent work~\cite{yu2015multi} proposed the use of dilated convolutions
for propagating multi-scale contextual information among CNN units.

A contribution of this work is to define convolutions over superpixels by defining
connectivity among them. In~\cite{he2015supercnn}, a method to use superpixels
inside CNNs has been proposed by re-arranging superpixels based on their
features. The technique proposed here is more generic and alleviates the need
for rearranging superpixels. A method to filter irregularly sampled data has been developed
in~\cite{bruna2013spectral} which may be applicable to superpixel convolutions. The difference
being that their method requires a pre-defined graph structure for every example/image
separately while our approach directly works on superpixels.
We experimented with Isomap embeddings~\cite{tenenbaum2000global} of superpixels but for speed reasons
opted for the more efficient kernels presented in this paper. The work of~\cite{mostajabi2014feedforward}
extracted multi-scale features at each superpixel and perform semantic segmentation
by classifying each superpixel independently. In contrast, we propagate
information across superpixels by using bilateral filters with learned feature
spaces.

Another core contribution of this work is the end-to-end trained bilateral filtering
module.
Several recent works on bilateral filtering~\cite{barron2015fast,barron2015defocus,kiefel15bnn,arxivpaper} back-propagate through permutohedral lattice approximation~\cite{adams2010fast}, to either learn the
filter parameters~\cite{kiefel15bnn,arxivpaper} or do optimization in the
bilateral space~\cite{barron2015fast,barron2015defocus}. Most of the existing works
on bilateral filtering use pre-defined feature spaces. In~\cite{campbell2013fully}, the feature
spaces for bilateral filtering are obtained via a non-parametric embedding into
an Euclidean space. In contrast, by explicitly computing the bilateral filter kernel,
we are able to back-propagate through features, thereby learning
the task-specific feature spaces for bilateral filters through integration into
end-to-end trainable CNNs.


\section{Superpixel Convolutional Networks}

We first formally introduce superpixels in Sec.~\ref{sec:superpixels} before we
describe the bilateral inception modules in Sec.~\ref{sec:inception}.

\subsection{Superpixels}\label{sec:superpixels}

The term \emph{superpixel} refers to a set of $n_i$ pixels $S_i=\{t_1,\ldots,t_{n_i}\}$ with
$t_k\in\{1,\ldots,N\}$ pixels. We use a set of $M$ superpixels $S=\{S_1,\ldots,S_M\}$
that are disjoint $S_i\cap S_j=\emptyset, \forall i,j$ and decompose the image,
$\cup_i S_i = \mathcal{I}$.

Superpixels have long been used for image segmentation in many previous
works, \eg~\cite{Gould:ECCV2014,gonfaus2010harmony,nowozin2010parameter,mostajabi2014feedforward},
as they provide a reduction of the problem size.
Instead of predicting a label $y_i$ for every pixel $x_i$, the classifier predicts a label $y_i$
per superpixel $S_i$ and extends this label to all pixels within.
A superpixel algorithm can pre-group pixels based on spatial and photometric similarity,
reducing the number of elements and also thereby regularizing the problem in a
meaningful way. The downside is that superpixels introduce a quantization error whenever
pixels within one segment have different ground truth label assignments.

\begin{wrapfigure}[18]{r}{5.3cm}
\includegraphics[width=0.4\textwidth]{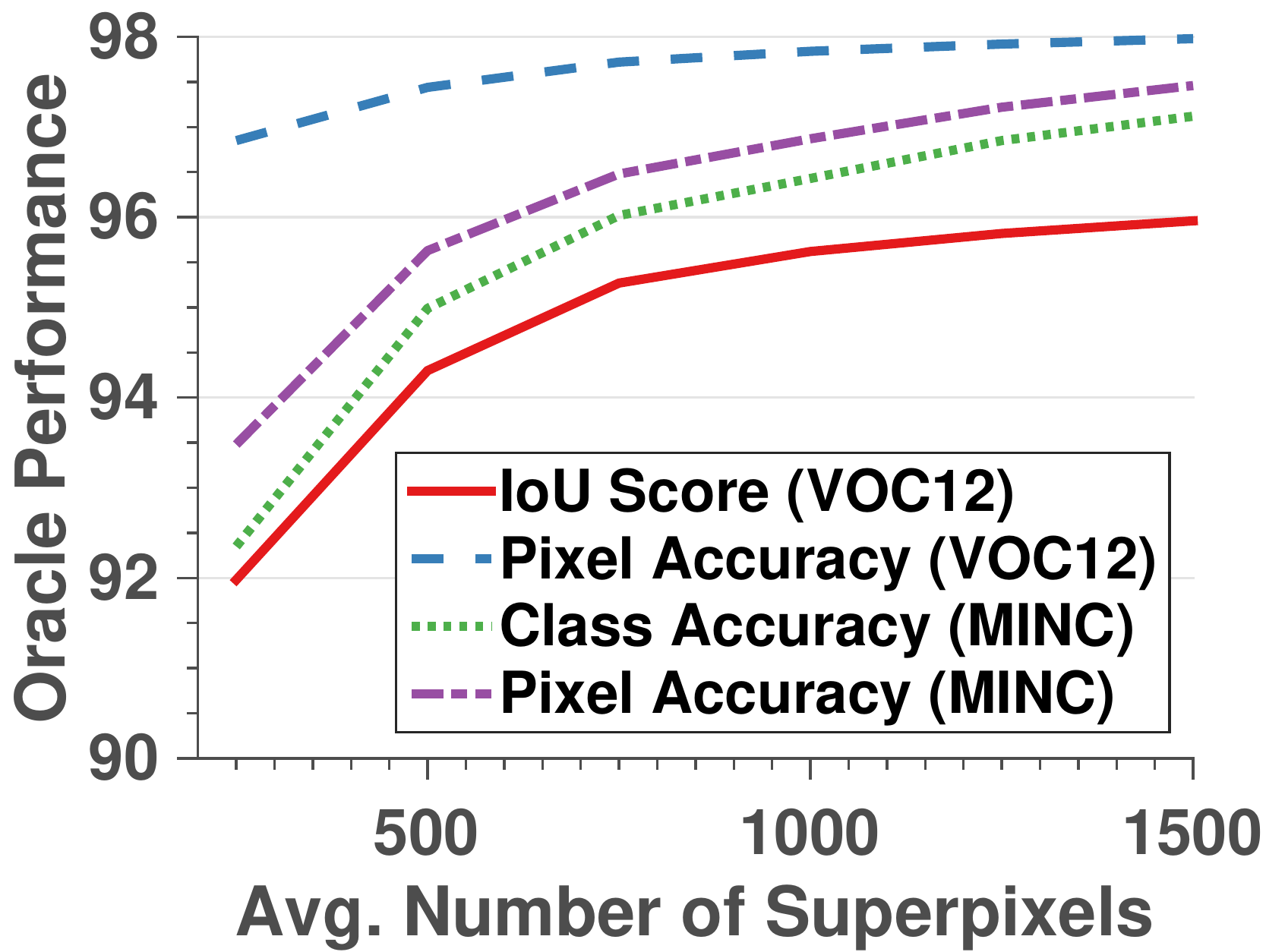}
  \mycaption{Superpixel Quantization Error} {Best achievable segmentation
    performance with a varying number
    of superpixels on Pascal VOC12 segmentation~\cite{voc2012segmentation} and MINC material
    segmentation~\cite{bell2015minc} datasets.}\label{fig:quantization}
\end{wrapfigure}

Figure~\ref{fig:quantization} shows the superpixel quantization effect with
the best achievable performance
as a function in the number of superpixels, on two different segmentation datasets:
PascalVOC~\cite{voc2012segmentation} and Materials in Context~\cite{bell2015minc}.
We find that the quantization effect is small compared to the current best segmentation
performance. Practically, we use the SLIC superpixels~\cite{achanta2012slic} for their runtime and~\cite{DollarICCV13edges} for their lower quantization error to decompose the image into superpixels. For details of the algorithms, please refer to the respective papers. We use publicly-available
real-time GPU implementation of SLIC, called gSLICr~\cite{gSLICr_2015}, which
runs at over 250Hz per second. And the publicly available Dollar superpixels code~\cite{DollarICCV13edges} computes a super-pixelization for a $400\times 500$ image in about 300ms using an Intel Xeon 3.33GHz CPU.

\subsection{Bilateral Inceptions}\label{sec:inception}

Next, we describe the \emph{Bilateral Inception Module} (BI) that performs Gaussian Bilateral
Filtering on multiple scales of the representations within a CNN.
The BI module can be inserted in between layers of existing CNN architectures.

\begin{figure}[t]
\begin{center}
\centerline{\includegraphics[width=0.9\textwidth]{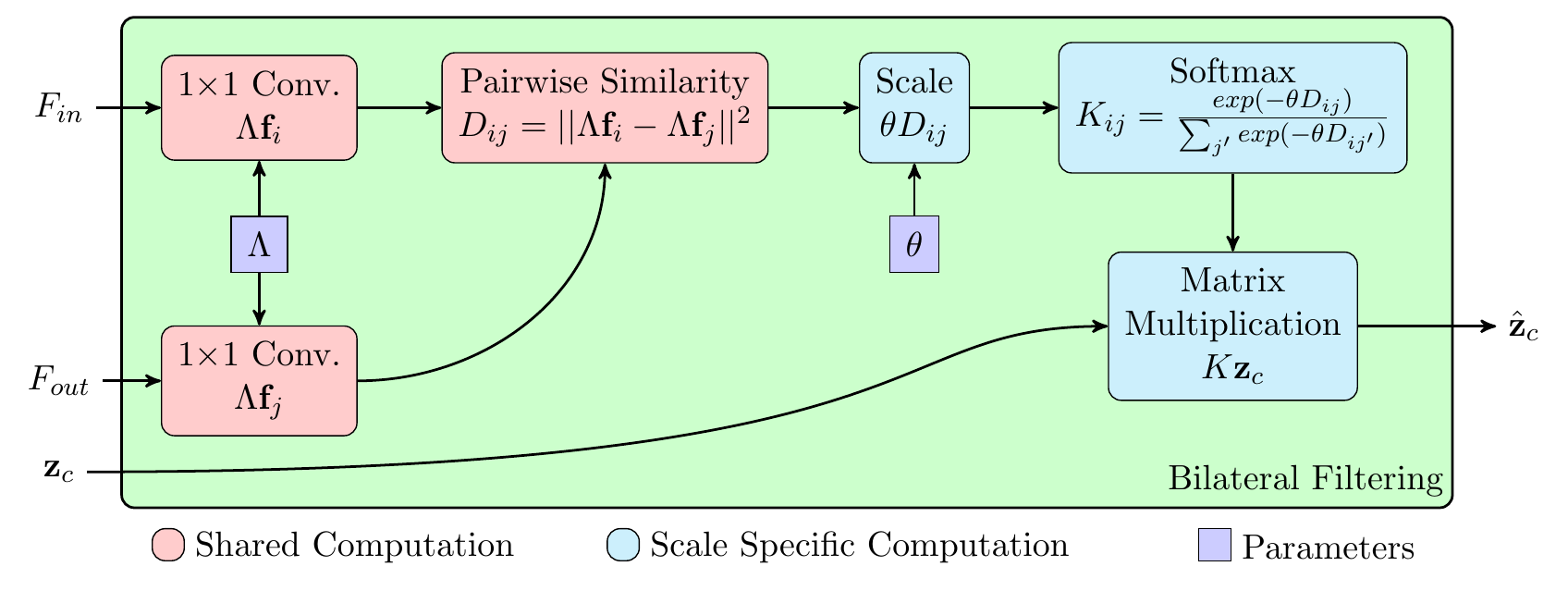}}
  \mycaption{Computation flow of the Gaussian Bilateral Filtering} {
We implemented the bilateral convolution with five separate computation blocks. $\Lambda$ and $\theta$
are the free parameters.}\label{fig:bi_module}
\end{center}
\end{figure}

{\bfseries Bilateral Filtering:} We first describe the Gaussian bilateral filtering,
the building block of the BI module.
A visualisation of the necessary computations is shown in Fig.~\ref{fig:bi_module}.
Given the previous layer CNN activations $\bz\in\mathbb{R}^{P\times C}$, that is $P$ points and $C$ filter responses.
With $\bz_c\in\mathbb{R}^P$ we denote the vector of activations of filter $c$.
Additionally we have for every point $j$ a feature vector $\bff_j\in\mathbb{R}^D$.
This denotes its spatial position ($D=2$, not necessarily a grid), position and RGB color ($D=5$), or others.
Separate from the input points with features $F_{in}=\{\bff_1,\ldots,\bff_P\}$ we have $Q$ output points with features $F_{out}$.
These can be the same set of points, but also fewer ($Q<P$), equal ($Q=P$), or more ($Q>P$) points.
For example we can filter a $10\times 10$ grid ($P=100$) and produce the result on a $50\times 50$ grid ($Q=2500$) or vice versa.

The bilateral filtered result will be denoted as $\hat{\bz}\in\mathbb{R}^{Q\times C}$.
We apply the same Gaussian bilateral filter to every channel $c$ separately.
A filter has two free parameters: the filter specific scale $\theta\in\mathbb{R}_+$ and the
global feature transformation parameters $\Lambda\in\mathbb{R}^{D\times D}$.
For $\Lambda$, a more general scaling could be applied using more features or a separate CNN.
Technically the bilateral filtering amounts to a matrix-vector multiplication $\forall c$:

\begin{equation}
\hat{\bz}_c = K(\theta, \Lambda, F_{in}, F_{out}) \bz_c,
\label{eq:filter}
\end{equation}
where $K\in\mathbb{R}^{Q\times P}$ and values for $f_i\in F_{out}, f_j\in F_{in}$:
\begin{equation}
K_{i,j} = \frac{\exp(-\theta\|\Lambda \bff_i- \Lambda \bff_j\|^2)}{\sum_{j'}\exp(-\theta\|\Lambda \bff_i- \Lambda \bff_{j'}\|^2)}.
\label{eq:filter}
\end{equation}

From a kernel learning terminology, $K$ is nothing but a Gaussian Gram matrix and it is symmetric if $F_{in}=F_{out}$.
We implemented this filtering in Caffe~\cite{jia2014caffe} using different layers as depicted in Fig.~\ref{fig:bi_module}.
While approximate computations of $K\bz_c$ exist and have improved runtime~\cite{adams2010fast,paris2006fast,gastal2011domain,adams2009gaussian}, we chose an explicit computation of $K$ due to its small size.
Our implementation makes use of GPU and the intermediate pairwise similarity computations are re-used across different modules.
The entire runtime is only a fraction of the CNN runtime, but of course applications to larger values of $P$ and $Q$ would require aforementioned algorithmic speed-ups.

\begin{figure}[t]
\begin{center}
\centerline{\includegraphics[width=0.9\textwidth]{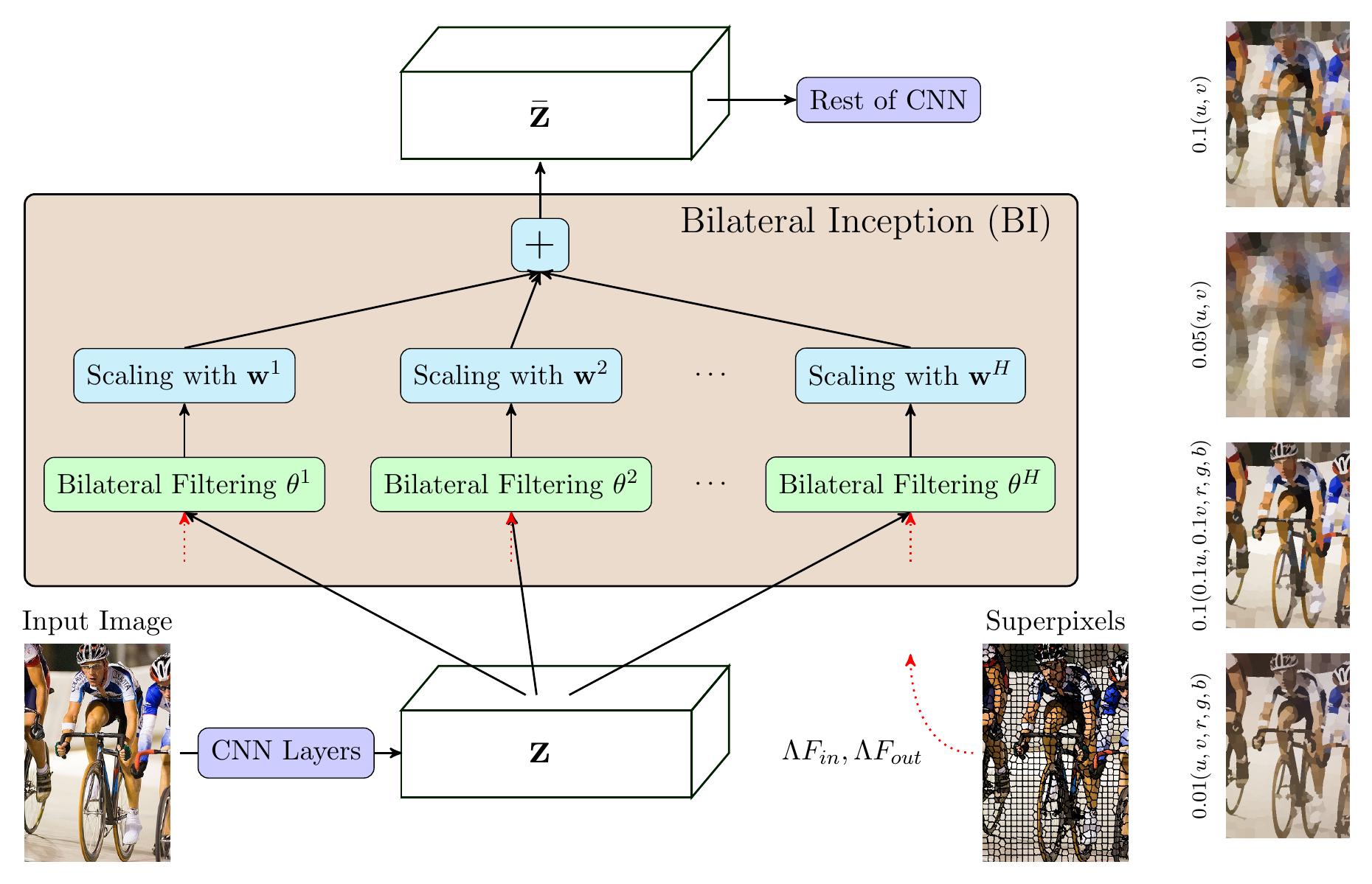}}
  \mycaption{Visualization of a Bilateral Inception (BI) Module} {The unit activations $\bz$
    are passed through several bilateral filters defined over different feature spaces. The result
  is linearly combined to $\bar{\bz}$ and passed on to the next network layer. Also shown are sample
  filtered superpixel images using bilateral filters defined over different example feature spaces. $(u,v)$ correspond to position and $(r,g,b)$ correspond to color features.}\label{fig:inception}
\end{center}
\end{figure}

{\bfseries Bilateral Inception Module:}
The \textit{bilateral inception module} (BI) is a weighted combination of different bilateral filters.
We combine the output of $H$ different filter kernels $K$, with different scales $\theta^1,\ldots,\theta^H$.
All kernels use the same feature transformation $\Lambda$ which allows for easier pre-computation of pairwise difference and avoids an over-parametrization of the filters.
The outputs of different filters $\hat{\bz}^h$ are combined linearly to produce $\bar{\bz}$:
\begin{equation}
\bar{\bz}_c = \sum_{h=1}^H \bw_c^h \hat{\bz}_c^h,
  \label{eq:module}
\end{equation}
using individual weights $\bw_c^h$ per scale $\theta^h$ and channel $c$.
The weights $\bw \in \mathbb{R}^{H\times C}$ are learned using error-backpropagation.
The result of the inception module has $C$ channels for every of its $Q$ points, thus $\bar{\bz} \in \mathbb{R}^{Q \times C}$.
The inception module is schematically illustrated in Fig.~\ref{fig:inception}.
In short, information from CNN layers below is filtered using bilateral filters defined in transformed feature space ($\Lambda \mathbf{f}$).
Most operations in the inception module are parallelizable resulting in fast runtimes on a GPU.
In this work, inspired from the DenseCRF architecture from~\cite{krahenbuhl2012efficient},
we used pairs of BI modules: one with position features $(u,v)$ and another with both position and
colour features $(u,v,r,g,b)$, each with multiple scales $\{\theta^h\}$.

{\bfseries Motivation and Comparison to DenseCRF:}
A BI module filters the activations of a CNN layer. Contrast this with the use of a DenseCRF on the CNN
output. At that point the fine-grained information that intermediate CNN layers represent has been condensed already to a low-dimensional vector representing beliefs over labels. Using a mean-field update is propagating information between these beliefs. Similar behaviour is obtained using the BI modules but on different scales (using multiple different filters $K(\theta^h)$) and on the intermediate CNN activations $\bz$. Since in the end, the to-be-predicted pixels are not i.i.d., this blurring leads to better performance both when using a bilateral filter as an approximate message passing step of a DenseCRF as well in the system outlined here. Both attempts are encoding prior knowledge about the problem, namely that pixels close in position and color are likely to have the same label. Therefore such pixels can also have the same intermediate representation. Consider one would average CNN representations for all pixels that have the same ground truth label. This would result in an intermediate CNN representation that would be very easy to classify for the later layers.

\subsection{Superpixel Convolutions}
The bilateral inception module allows to change how information is stored in the higher level of a CNN. This is where the superpixels are used.
Instead of storing information on a fixed grid, we compute for every image, superpixels $S$ and use the mean color and position of their included pixels as features.
We can insert bilateral inception modules to change from grid representations to superpixel representations and vice versa.
Inception modules in between superpixel layers convolve the unit activations between all superpixels depending on their distance in the feature space.
This retains all properties of the bilateral filter, superpixels that are spatially close and have a similar mean color will have a stronger influence on each other.

Superpixels are not the only choice, in principle one can also sample random points from the image and use them as intermediate representations.
We are using superpixels for computational reasons, since they can be used to propagate label information to the full image resolution.
Other interpolation techniques are possible, including the well known bilinear interpolation, up-convolution networks~\cite{zeiler2010deconvolutional}, and DenseCRFs~\cite{krahenbuhl2012efficient}.
The quantization error mentioned in Sec.~\ref{sec:superpixels} only enters because the superpixels are used for interpolation.
Also note that a fixed grid, that is independent of the image is a hard choice of where information should be stored.
One could in principle evaluate the CNN densely, at all possible spatial locations,
but we found that this resulted in poor performance compared to interpolation methods.

\subsubsection{Backpropagation and Training.}

All free parameters of the inception module $\bw$, $\{\theta^h\}$ and $\Lambda$ are learned
via backpropagation. We also backpropagate the error with respect to the module inputs
thereby enabling the integration of our inception modules inside CNN frameworks without
breaking the end-to-end learning paradigm.
As shown in Fig.~\ref{fig:bi_module}, the bilateral filtering
can be decomposed into 5 different sub-layers.
Derivatives with respect to the open parameters are obtained by the
corresponding layer and standard backpropagation through the directed
acyclic graph.
For example, $\Lambda$ is optimized by back-propagating gradients through $1\times1$ convolution.
Derivatives for non-standard layers (pairwise similarity, matrix
multiplication) are straight forward to obtain using matrix calculus.
To let different filters learn the information propagation at
different scales, we initialized $\{\theta^h\}$ with well separated
scalar values (\eg $\{1, 0.7, 0.3,...\}$).
The learning is performed using Adam stochastic optimization
method~\cite{kingma2014adam}.
The implementation is done in Caffe neural network framework~\cite{jia2014caffe},
and the code is available online at http://segmentation.is.tuebingen.mpg.de.

\definecolor{voc_1}{RGB}{0, 0, 0}
\definecolor{voc_2}{RGB}{128, 0, 0}
\definecolor{voc_3}{RGB}{0, 128, 0}
\definecolor{voc_4}{RGB}{128, 128, 0}
\definecolor{voc_5}{RGB}{0, 0, 128}
\definecolor{voc_6}{RGB}{128, 0, 128}
\definecolor{voc_7}{RGB}{0, 128, 128}
\definecolor{voc_8}{RGB}{128, 128, 128}
\definecolor{voc_9}{RGB}{64, 0, 0}
\definecolor{voc_10}{RGB}{192, 0, 0}
\definecolor{voc_11}{RGB}{64, 128, 0}
\definecolor{voc_12}{RGB}{192, 128, 0}
\definecolor{voc_13}{RGB}{64, 0, 128}
\definecolor{voc_14}{RGB}{192, 0, 128}
\definecolor{voc_15}{RGB}{64, 128, 128}
\definecolor{voc_16}{RGB}{192, 128, 128}
\definecolor{voc_17}{RGB}{0, 64, 0}
\definecolor{voc_18}{RGB}{128, 64, 0}
\definecolor{voc_19}{RGB}{0, 192, 0}
\definecolor{voc_20}{RGB}{128, 192, 0}
\definecolor{voc_21}{RGB}{0, 64, 128}
\definecolor{voc_22}{RGB}{128, 64, 128}

\begin{figure*}[t]
  \tiny
  \centering

  \subfigure{%
    \includegraphics[width=.15\columnwidth]{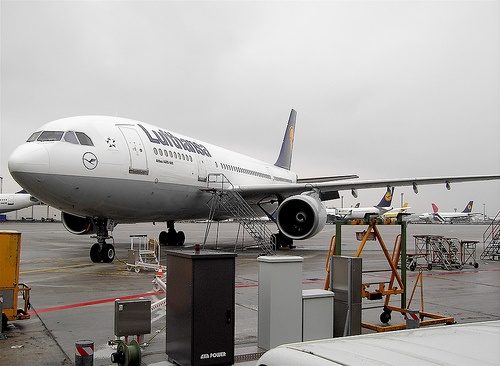}
  }
  \subfigure{%
    \includegraphics[width=.15\columnwidth]{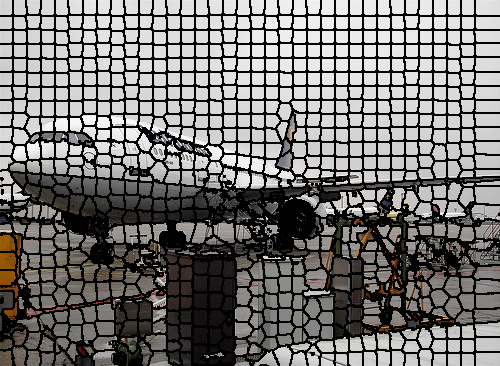}
  }
  \subfigure{%
    \includegraphics[width=.15\columnwidth]{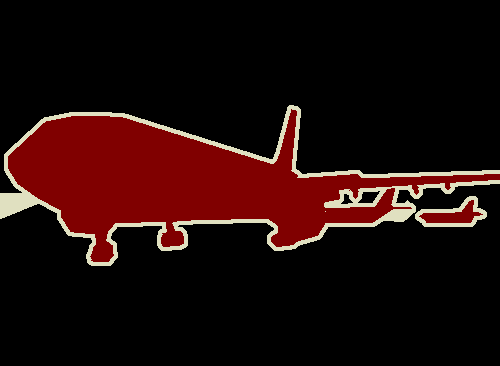}
  }
  \subfigure{%
    \includegraphics[width=.15\columnwidth]{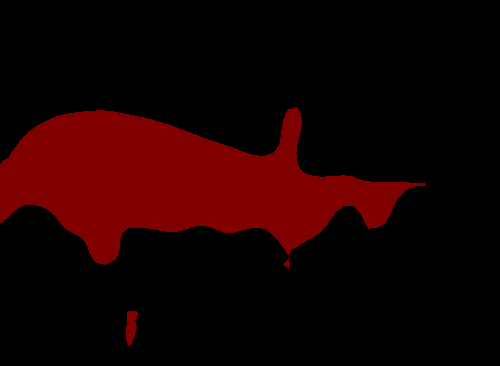}
  }
  \subfigure{%
    \includegraphics[width=.15\columnwidth]{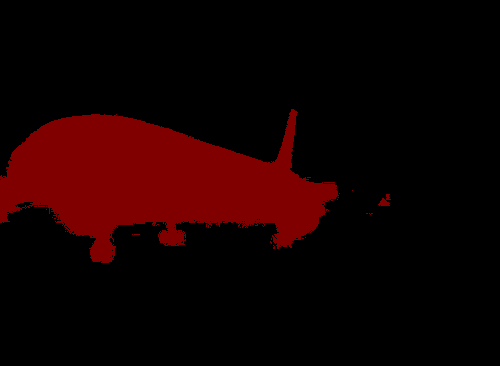}
  }
  \subfigure{%
    \includegraphics[width=.15\columnwidth]{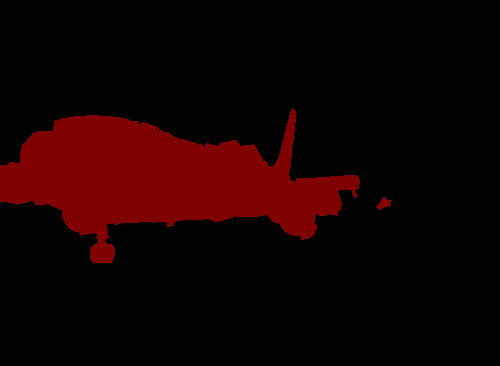}
  }\\[-2ex]
  \setcounter{subfigure}{0}
  \subfigure[\tiny Input]{%
    \includegraphics[width=.15\columnwidth]{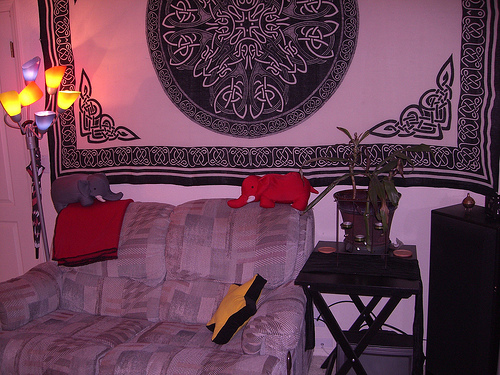}
  }
  \subfigure[\tiny Superpixels]{%
    \includegraphics[width=.15\columnwidth]{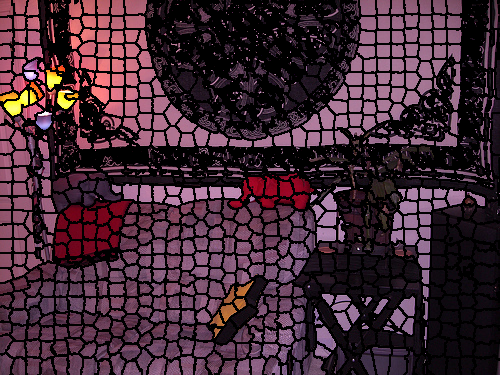}
  }
  \subfigure[\tiny GT]{%
    \includegraphics[width=.15\columnwidth]{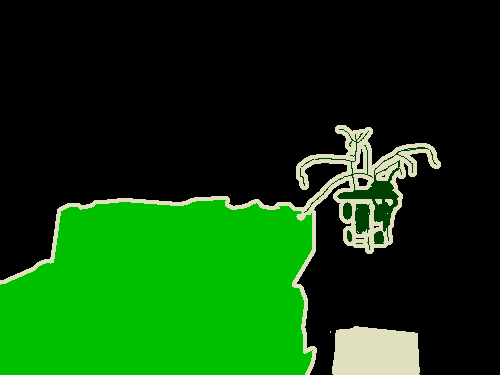}
  }
  \subfigure[\tiny Deeplab]{%
    \includegraphics[width=.15\columnwidth]{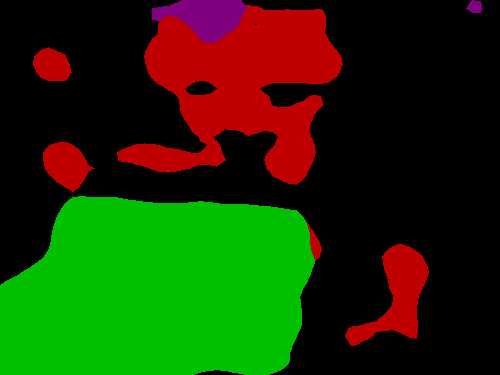}
  }
  \subfigure[\tiny +DenseCRF]{%
    \includegraphics[width=.15\columnwidth]{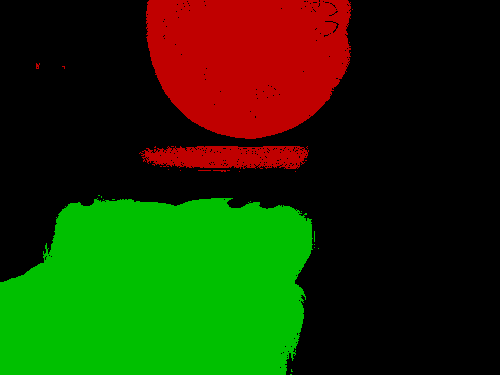}
  }
  \subfigure[\tiny Using BI]{%
    \includegraphics[width=.15\columnwidth]{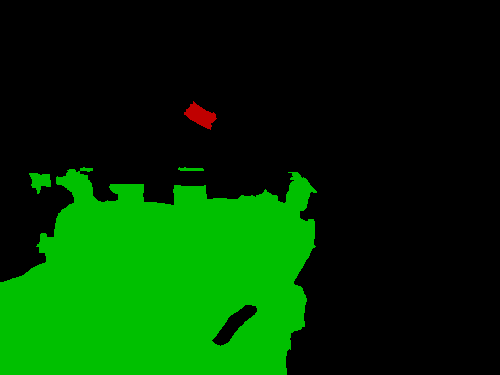}
  }
  \mycaption{Semantic Segmentation}{Example results of semantic segmentation
  on Pascal VOC12 dataset.
  (d) depicts the DeepLab CNN result, (e) CNN + 10 steps of mean-field inference,
  (f) result obtained with bilateral inception (BI) modules (\bi{6}{2}+\bi{7}{6}) between \fc~layers.}\label{fig:semantic_visuals}
\end{figure*}

\section{Experiments}

We study the effect of inserting and learning bilateral inception modules
in various existing CNN architectures.
As a testbed we perform experiments on semantic segmentation
using the Pascal VOC2012 segmentation
benchmark dataset~\cite{voc2012segmentation}, Cityscapes street scene dataset~\cite{Cordts2015Cvprw}
and on material segmentation using the Materials in Context (MINC)
dataset from~\cite{bell2015minc}. We take different CNN architectures from the works
of~\cite{chen2014semantic,zheng2015conditional,bell2015minc} and insert the
inception modules before and/or after the spatial FC layers. In the supplementary, we presented
some quantitative results with approximate bilateral filtering using the permutohedral lattice~\cite{adams2010fast}.

\subsection{Semantic Segmentation}

We first use the Pascal VOC12 segmentation dataset~\cite{voc2012segmentation}
with 21 object classes.
For all experiments on VOC2012, we train using the extended training set of
10581 images collected by~\cite{hariharan2011moredata}. Following~\cite{zheng2015conditional},
we use a reduced validation set of 346 images for validation.
We experiment on two different network architectures, (a) DeepLab model from~\cite{chen2014semantic} which uses CNN followed by DenseCRF and (b) CRFasRNN model from~\cite{zheng2015conditional} which uses
CNN with deconvolution layers followed by DenseCRF trained end-to-end.

\subsubsection{DeepLab}\label{sec:deeplabmodel}
We use the publicly available state-of-the-art pre-trained CNN models from~\cite{chen2014semantic}.
We use the DeepLab-LargeFOV variant as a base architecture and refer to it as `\deeplab'.
The \deeplab~CNN model produces a lower resolution prediction ($\frac{1}{8}\times$)
which is then bilinearly interpolated to the input image resolution.
The original models have been fine-tuned using both the MSCOCO~\cite{lin2014microsoft} and the extended VOC~\cite{hariharan2011moredata} datasets.
Next, we describe modifications to these models and show performance improvements in terms of both IoU and runtimes.

\begin{wraptable}[24]{r}{0pt}
  \scriptsize
  \begin{tabular}{>{\raggedright\arraybackslash}p{3.3cm}>{\raggedright\arraybackslash}p{1.2cm}>{\centering\arraybackslash}p{0.8cm}>{\centering\arraybackslash}p{1.0cm}}
    \toprule
    \textbf{Model} & Training & \emph{IoU} & \emph{Runtime}\\
    \midrule
    \scriptsize
  \deeplablargefov~\cite{chen2014semantic} &  & 68.9 & 145ms\\
  \midrule
  With BI modules & & & \\
  \bi{6}{2} & only BI & \href{http://host.robots.ox.ac.uk:8080/anonymous/31URIG.html}{70.8} & +20 \\
  \bi{6}{2} & BI+FC & \href{http://host.robots.ox.ac.uk:8080/anonymous/JOB8CE.html}{71.5} & +20\\
  \bi{6}{6} & BI+FC & \href{http://host.robots.ox.ac.uk:8080/anonymous/IB1UAZ.html}{72.9} & +45\\
  \bi{7}{6} & BI+FC & \href{http://host.robots.ox.ac.uk:8080/anonymous/EQB3CR.html}{73.1} & +50\\
  \bi{8}{10} & BI+FC & \href{http://host.robots.ox.ac.uk:8080/anonymous/JR27XL.html}{72.0} & +30\\
  \bi{6}{2}-\bi{7}{6} & BI+FC & \href{http://host.robots.ox.ac.uk:8080/anonymous/VOTV5E.html}{73.6} & +35\\
  \bi{7}{6}-\bi{8}{10} & BI+FC & \href{http://host.robots.ox.ac.uk:8080/anonymous/X7A3GP.html}{73.4} & +55\\
  \bi{6}{2}-\bi{7}{6} & FULL &  \href{http://host.robots.ox.ac.uk:8080/anonymous/CLLB3J.html}{\textbf{74.1}} & +35\\
  \bi{6}{2}-\bi{7}{6}-CRF & FULL & \href{http://host.robots.ox.ac.uk:8080/anonymous/7NGWWU.html}{\textbf{75.1}} & +865\\
  \midrule
  \deeplablargefovcrf~\cite{chen2014semantic} & & 72.7 & +830\\
  \deeplabmsclargefovcrf~\cite{chen2014semantic} & & \textbf{73.6} & +880\\
  DeepLab-EdgeNet~\cite{chen2015semantic} & & 71.7 & +30\\
  DeepLab-EdgeNet-CRF~\cite{chen2015semantic} & & \textbf{73.6} & +860\\
  \bottomrule
\end{tabular}
\mycaption{Semantic Segmentation using \deeplablargefov~model}
{IoU scores on Pascal VOC12 segmentation test dataset
and average runtimes (ms) corresponding to different models. Also shown are the results
corresponding to competitive dense pixel prediction techniques that used
the same base DeepLab CNN. Runtimes also include superpixel
computation (6ms). In the second column, `BI', `FC' and `FULL'
correspond to training `BI', `FC' and full model layers respectively.}
\label{tab:deeplabresults}
\end{wraptable}

We add inception modules after different FC layers in the original model and
remove the DenseCRF post processing. For this dataset, we use
1000 SLIC superpixels~\cite{achanta2012slic,gSLICr_2015}.
The inception modules after \fc{6}, \fc{7} and \fc{8} layers are referred to as
\bi{6}{H}, \bi{7}{H} and \bi{8}{H} respectively, where $H$ is the number
of kernels. All results using the~\deeplab~model on Pascal VOC12 dataset
are summarized in Tab.~\ref{tab:deeplabresults}.
We report the `test' numbers without validation numbers, because the released
DeepLab model that we adapted was trained using both train and validation sets.
The~\deeplab~network achieves an IoU of 68.9 after bilinear interpolation.
Experiments with \bi{6}{2}
module indicate that even only learning the inception module while keeping the remaining
network fixed results in an reliable IoU improvement ($+1.9$).
Additional joint training with \fc{} layers significantly improved the performance.
The results also show that more kernels improve performance.
Next, we add multiple modules to the base DeepLab network at various stages and train them jointly. This
results in further improvement of the performance. The \bi{6}{2}-\bi{7}{6} model with
two inception modules shows significant improvement in IoU by $4.7$ and $0.9$
 in comparison to baseline model and DenseCRF application respectively.
Finally, finetuning the entire network (FULL in Tab.~\ref{tab:deeplabresults})
boosts the performance by $5.2$ and $1.4$ compared to the baseline and DenseCRF application.

Some visual results are shown in Fig.~\ref{fig:semantic_visuals} and more
are included in the supplementary.
Several other variants of using BI are conceivable. During our experiments,
we have observed that more kernels and more modules improve the performance,
so we expect that even better results can be achieved.
In Tab.~\ref{tab:deeplabresults}, the runtime (ms) is included for several models.
These numbers have been obtained using a
Nvidia Tesla K80 GPU and standard Caffe time benchmarking~\cite{jia2014caffe}.
DenseCRF timings are taken from~\cite{chen2015semantic}. The runtimes indicate
that the overhead with BI modules is quite minimal in comparison to
using Dense CRF.

In addition, we include the results of some other dense pixel prediction methods that
are build on top of the same DeepLab base model. \deeplabmsclargefovcrf~is a multi-scale
version~\cite{chen2014semantic} of \deeplab~with DenseCRF on top. DeepLab-EdgeNet~\cite{chen2015semantic}
is a recently proposed fast and discriminatively trained domain transform technique
for propagating information across pixels. Comparison with these techniques in terms
of performance and runtime indicates that our approach performs on par with latest
dense pixel prediction techniques with significantly less time overhead.
Several
state-of-the-art CNN based systems~\cite{lin2015efficient,liu2015semantic}
have achieved higher results than \deeplab~on Pascal VOC12. These models are not yet
publicly available and so we could not test the use of BI models in them.
A close variant~\cite{barron2015fast} of our work, which propose
to do optimization in the bilateral space also has fast runtimes, but reported
lower performance in comparison to the application of DenseCRF.

\begin{wraptable}[14]{r}{0pt}
  \scriptsize
  \begin{tabular}{>{\raggedright\arraybackslash}p{4.0cm}>{\centering\arraybackslash}p{0.7cm}>{\centering\arraybackslash}p{1.0cm}}
    \toprule
    \textbf{Model} & \emph{IoU} & \emph{Runtime}\\
    \midrule
    \scriptsize
  \deconv(CNN+Deconv.) & 72.0 & 190ms \\
  \midrule
  With BI modules & & \\
  \bi{3}{2}-\bi{4}{2}-\bi{6}{2}-\bi{7}{2} & \textbf{74.9} & 245 \\
  \midrule
  CRFasRNN (\deconvcrf)& 74.7 & 2700\\
  \bottomrule
\end{tabular}
\mycaption{Semantic Segmentation using CRFasRNN model}{IoU scores and runtimes corresponding to different models on Pascal VOC12 test dataset. Note that runtime also includes superpixel computation.}
\label{tab:deconvresults}
\end{wraptable}

\subsubsection{CRFasRNN}
As a second architecture, we modified the CNN architecture
trained by~\cite{zheng2015conditional} that
produces a result at an even lower resolution ($\frac{1}{16} \times$).
Multiple deconvolution steps are employed to obtain the segmentation at input
image resolution. This result is then passed onto the DenseCRF recurrent neural network
to obtain the final segmentation result.
We insert BI modules after score-pool3, score-pool4, \fc{6} and \fc{7} layers, please
see~\cite{long2014fully,zheng2015conditional} for the network architecture details.
Instead of combining outputs from the above layers with deconvolution steps, we introduce
BI modules after them and linearly combined the outputs to obtain final segmentation result.
Note that we entirely removed both the deconvolution and the DenseCRF parts of the original model~\cite{zheng2015conditional}.
See Tab.~\ref{tab:deconvresults} for results on the DeconvNet model.
Without the DenseCRF part and only evaluating the deconvolutional part of this
model, one obtains an IoU score of $72.0$.
Ten steps of mean field inference increase the IoU to $74.7$~\cite{zheng2015conditional}. Our model, with few additional parameters compared to the base CNN,
achieves a IoU performance of $74.9$, showing an improvement of 0.2 over the CRFasRNN model.
The BI layers lead to better performance than deconvolution and DenseCRF combined while being much faster.

\subsubsection{Hierarchical Clustering Analysis}
\begin{figure}[t]
\begin{tabular}{c}
  \subfigure{%
    \includegraphics[width=0.25\textwidth]{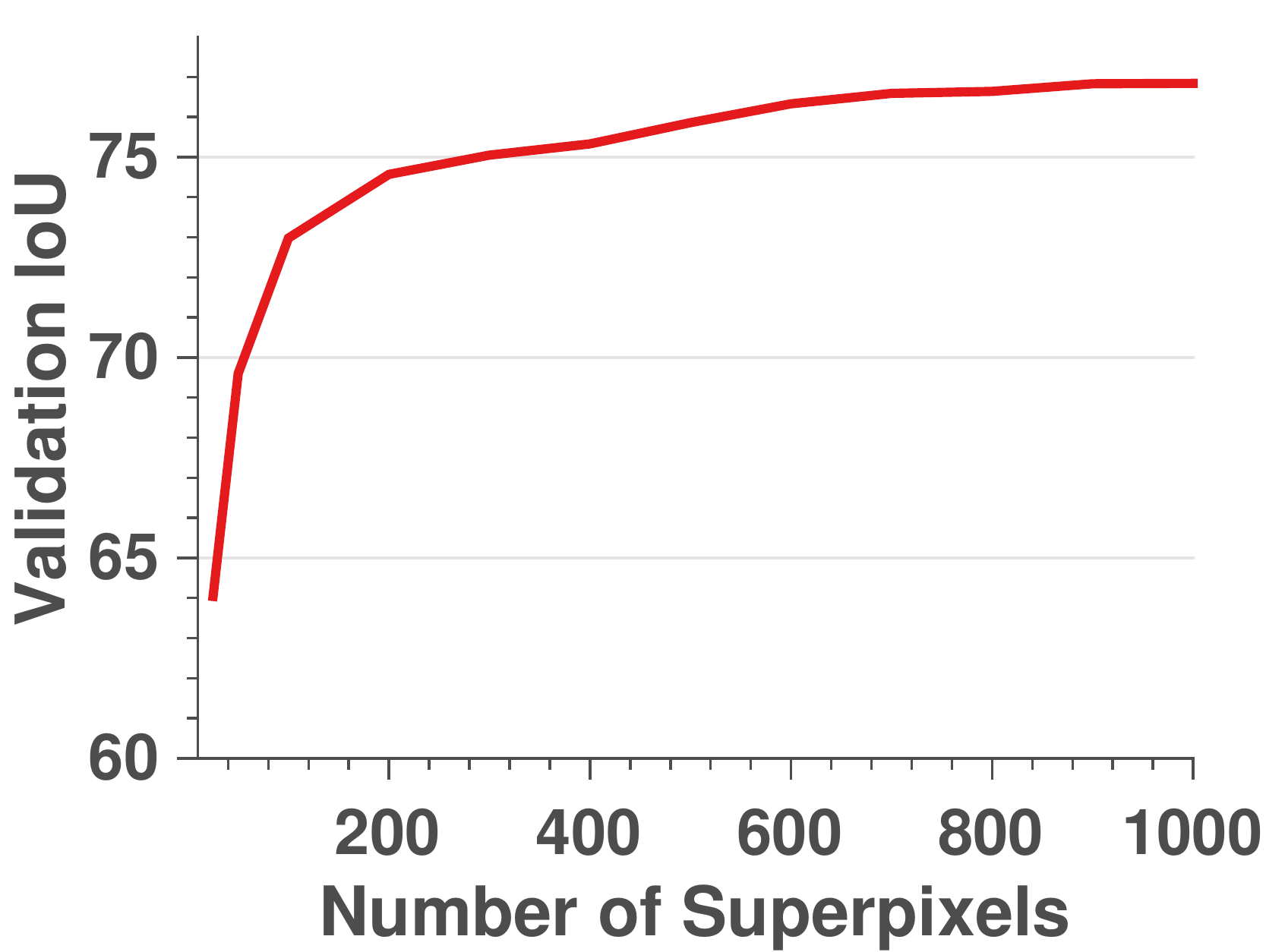}
  }
\end{tabular} \hfill
\begin{tabular}{c}
  \subfigure{%
    \includegraphics[width=0.14\textwidth]{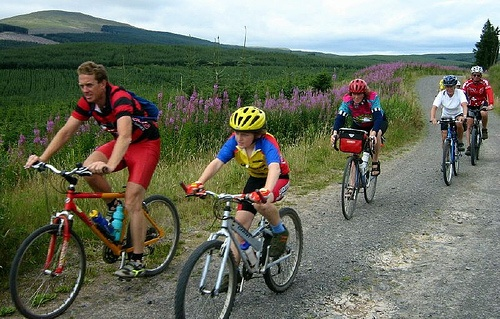}
  }\\
  \subfigure{%
    \includegraphics[width=0.14\textwidth]{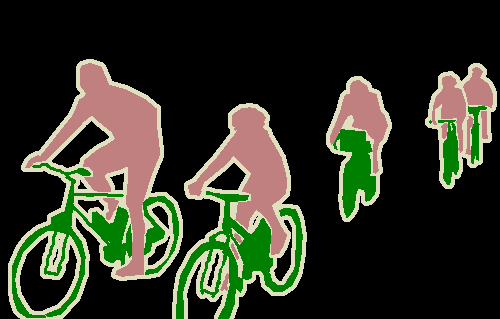}
  }
\end{tabular} \hfill
\begin{tabular}{c}
  \subfigure{%
    \includegraphics[width=0.14\textwidth]{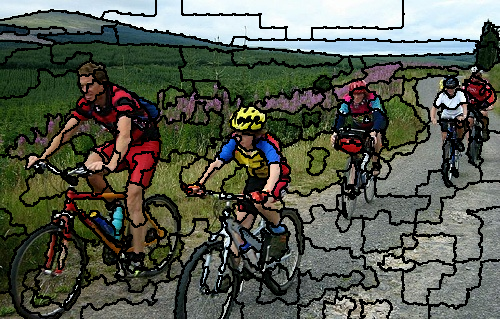}
  }\\
  \subfigure{%
    \includegraphics[width=0.14\textwidth]{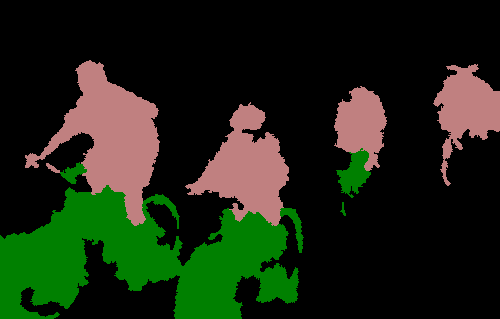}
  }
\end{tabular}\hfill
\begin{tabular}{c}
  \subfigure{%
    \includegraphics[width=0.14\textwidth]{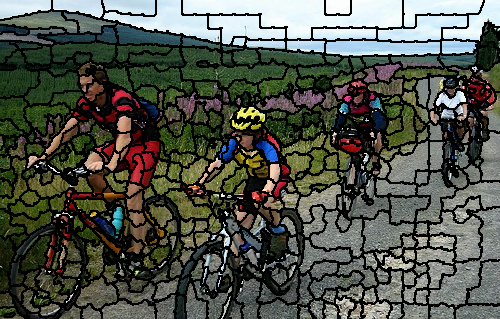}
  }\\
  \subfigure{%
    \includegraphics[width=0.14\textwidth]{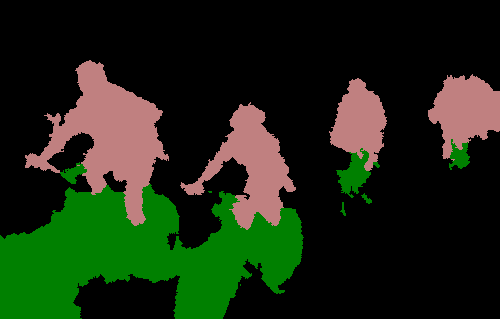}
  }
\end{tabular}\hfill
\begin{tabular}{c}
  \subfigure{%
    \includegraphics[width=0.14\textwidth]{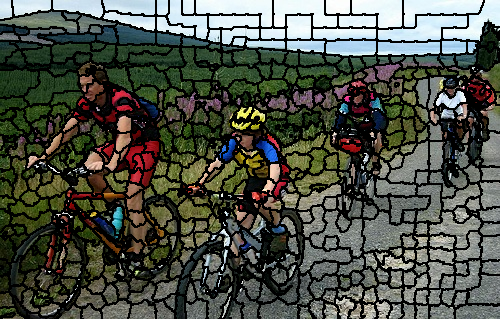}
  }\\
  \subfigure{%
    \includegraphics[width=0.14\textwidth]{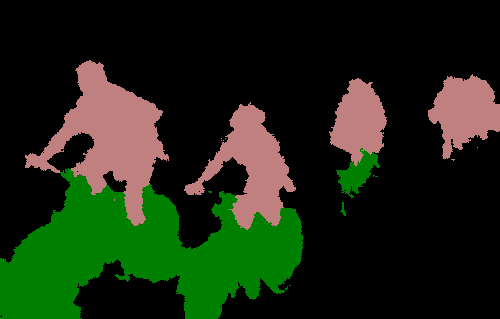}
  }
\end{tabular}
  \mycaption{Hierarchical Clustering Analysis}{From left to right: Validation performance when using different super-pixel layouts, visualization of an image with ground truth segmentation, and the \bi{6}{2}-\bi{7}{6} result with 200, 600, and 1000 superpixels.}
  \label{fig:clustering}
\end{figure}

We learned the network parameters using 1000 gSLIC superpixels per image, however the inception module allows to change the resolution (a non-square $K$).
To illustrate this, we perform agglomorative clustering of the superpixels, sequentially merging the nearest two superpixels into a single one.
We then evaluated the DeepLab-\bi{6}{2}-\bi{7}{6} network using different levels of the resulting hierarchy re-using all the trained network parameters.
Results in Fig.~\ref{fig:clustering} show that the IoU score on the validation set decreases slowly with decreasing number of points and then drops for less than 200 superpixels.
This validates that the network generalizes to different superpixel layouts and it is sufficient to represent larger regions of similar color by fewer points.
In future, we plan to explore different strategies to allocate the representation to those regions that require more resolution and to remove the superpixelization altogether.
Fig.~\ref{fig:clustering} shows example image with 200, 600, and 1000 superpixels and their obtained segmentation with BI modules.

\begin{wraptable}[15]{r}{0pt}
  \scriptsize
  \centering
  \begin{tabular}{p{2.2cm}>{\centering\arraybackslash}p{2cm}>{\centering\arraybackslash}p{1cm}}
    \toprule
    \textbf{Model} & Class / Total accuracy & Runtime\\
    \midrule
    \scriptsize
    Alexnet CNN & 55.3 / 58.9 & 300ms \\
    \midrule
    \bi{7}{2}-\bi{8}{6} & 67.7 / 71.3 & 410 \\
    \bi{7}{6}-\bi{8}{6} & \textbf{69.4 / 72.8} & 470 \\
    \midrule
    AlexNet-CRF & 65.5 / 71.0 & 3400 \\
    \bottomrule
  \end{tabular}
  \mycaption{Material Segmentation using AlexNet}{Pixel accuracies and runtimes (in ms)
  of different models on MINC material segmentation dataset~\cite{bell2015minc}.
  Runtimes also include the time for superpixel extraction (15ms).}
  \label{tab:mincresults}
\end{wraptable}

\subsection{Material Segmentation}

We also experiment on a different pixel prediction task of
material segmentation by adapting a CNN
architecture finetuned for Materials in Context (MINC)~\cite{bell2015minc} dataset.
MINC consists of 23 material classes and is available in three different
resolutions with the same aspect ratio: low ($550^2$), mid ($1100^2$) and an original higher
resolution.
The authors of~\cite{bell2015minc} train CNNs on the mid resolution images and then
combine with a DenseCRF to predict and evaluate on low resolution images.
We build our work based on the Alexnet model~\cite{krizhevsky2012imagenet}
released by the authors of~\cite{bell2015minc}. To obtain a
per pixel labeling of a given image, there are several
processing steps that~\cite{bell2015minc} use for good performance. First,
a CNN is applied at several scales with different strides followed by an interpolation of
the predictions to reach the input image resolution and is then followed by a
DenseCRF. For simplicity, we choose to run the CNN network
with single scale and no-sliding. The authors used just one kernel
with $(u, v, L, a, b)$ features in the DenseCRF part.
We used the same features in our inception modules.
We modified the base AlexNet model by inserting BI modules after \fc{7}
and \fc{8} layers. Again, 1000 SLIC superpixels are used for all experiments.
Results on the test set are shown in Table~\ref{tab:mincresults}.
When inserting BI modules, the performance improves both in total pixel accuracy as well as in class-averaged
accuracy. We observe an improvement
of $12\%$ compared to CNN predictions and $2-4\%$ compared to CNN+DenseCRF results.
Qualitative examples are shown in Fig.~\ref{fig:material_visuals} and more are included
in the supplementary. The weights to combine outputs in the BI layers are
found by validation on the validation set. For this model we do not provide any
learned setup due very limited segment training data.

\definecolor{minc_1}{HTML}{771111}
\definecolor{minc_2}{HTML}{CAC690}
\definecolor{minc_3}{HTML}{EEEEEE}
\definecolor{minc_4}{HTML}{7C8FA6}
\definecolor{minc_5}{HTML}{597D31}
\definecolor{minc_6}{HTML}{104410}
\definecolor{minc_7}{HTML}{BB819C}
\definecolor{minc_8}{HTML}{D0CE48}
\definecolor{minc_9}{HTML}{622745}
\definecolor{minc_10}{HTML}{666666}
\definecolor{minc_11}{HTML}{D54A31}
\definecolor{minc_12}{HTML}{101044}
\definecolor{minc_13}{HTML}{444126}
\definecolor{minc_14}{HTML}{75D646}
\definecolor{minc_15}{HTML}{DD4348}
\definecolor{minc_16}{HTML}{5C8577}
\definecolor{minc_17}{HTML}{C78472}
\definecolor{minc_18}{HTML}{75D6D0}
\definecolor{minc_19}{HTML}{5B4586}
\definecolor{minc_20}{HTML}{C04393}
\definecolor{minc_21}{HTML}{D69948}
\definecolor{minc_22}{HTML}{7370D8}
\definecolor{minc_23}{HTML}{7A3622}
\definecolor{minc_24}{HTML}{000000}

\begin{figure*}[t]
  \tiny 
  \centering
  \subfigure{%
    \includegraphics[width=.15\columnwidth]{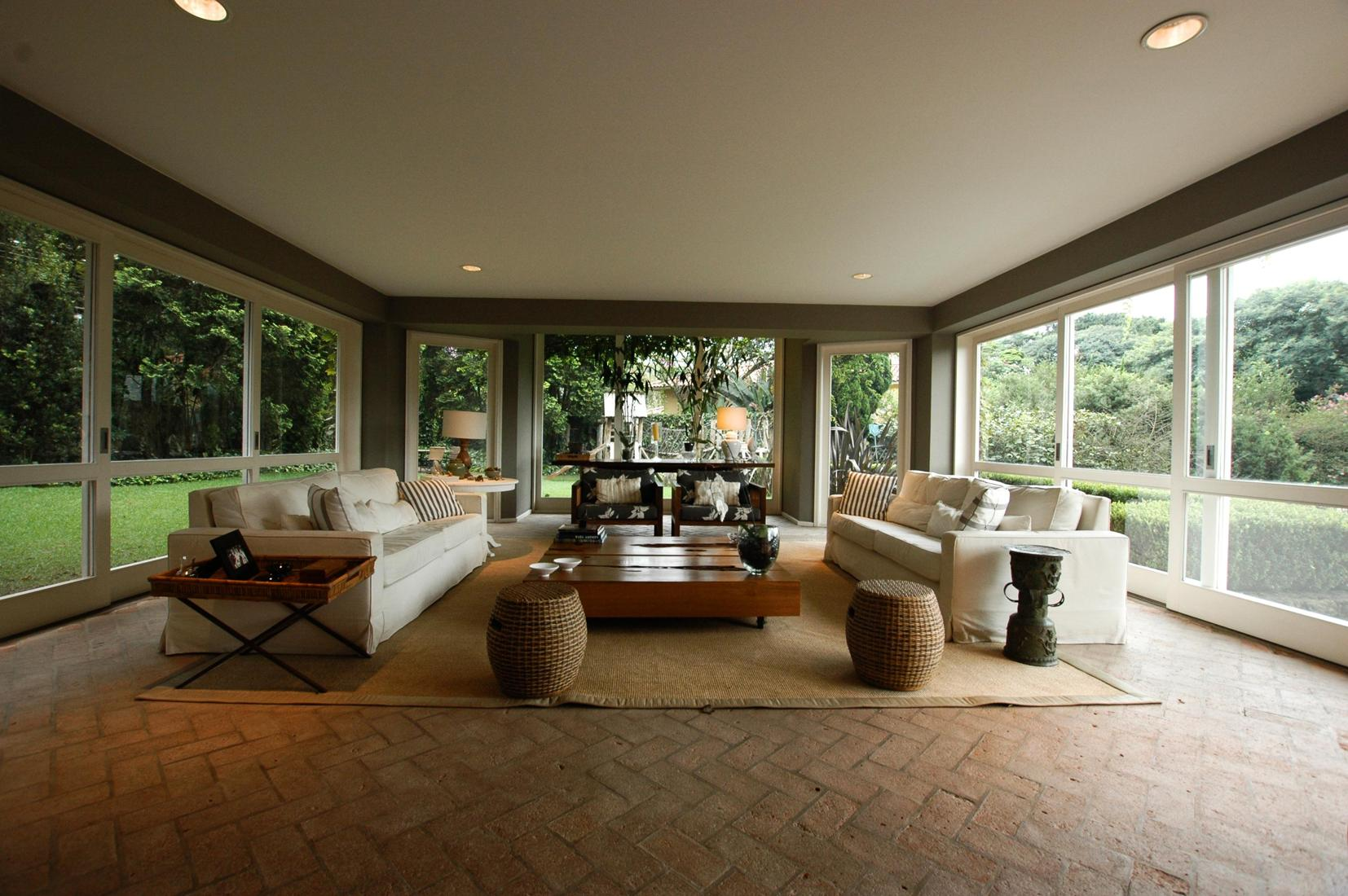}
  }
  \subfigure{%
    \includegraphics[width=.15\columnwidth]{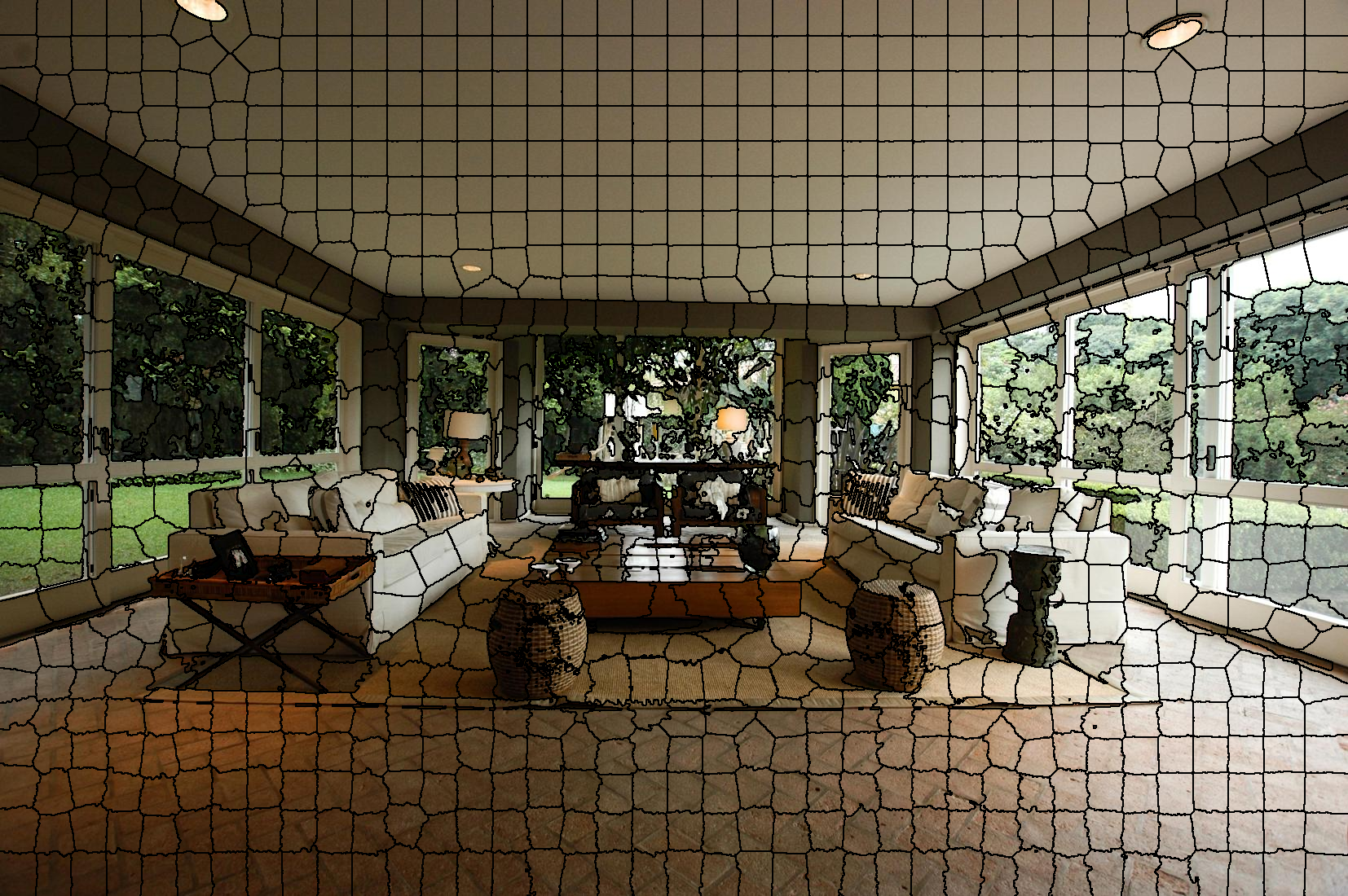}
  }
  \subfigure{%
    \includegraphics[width=.15\columnwidth]{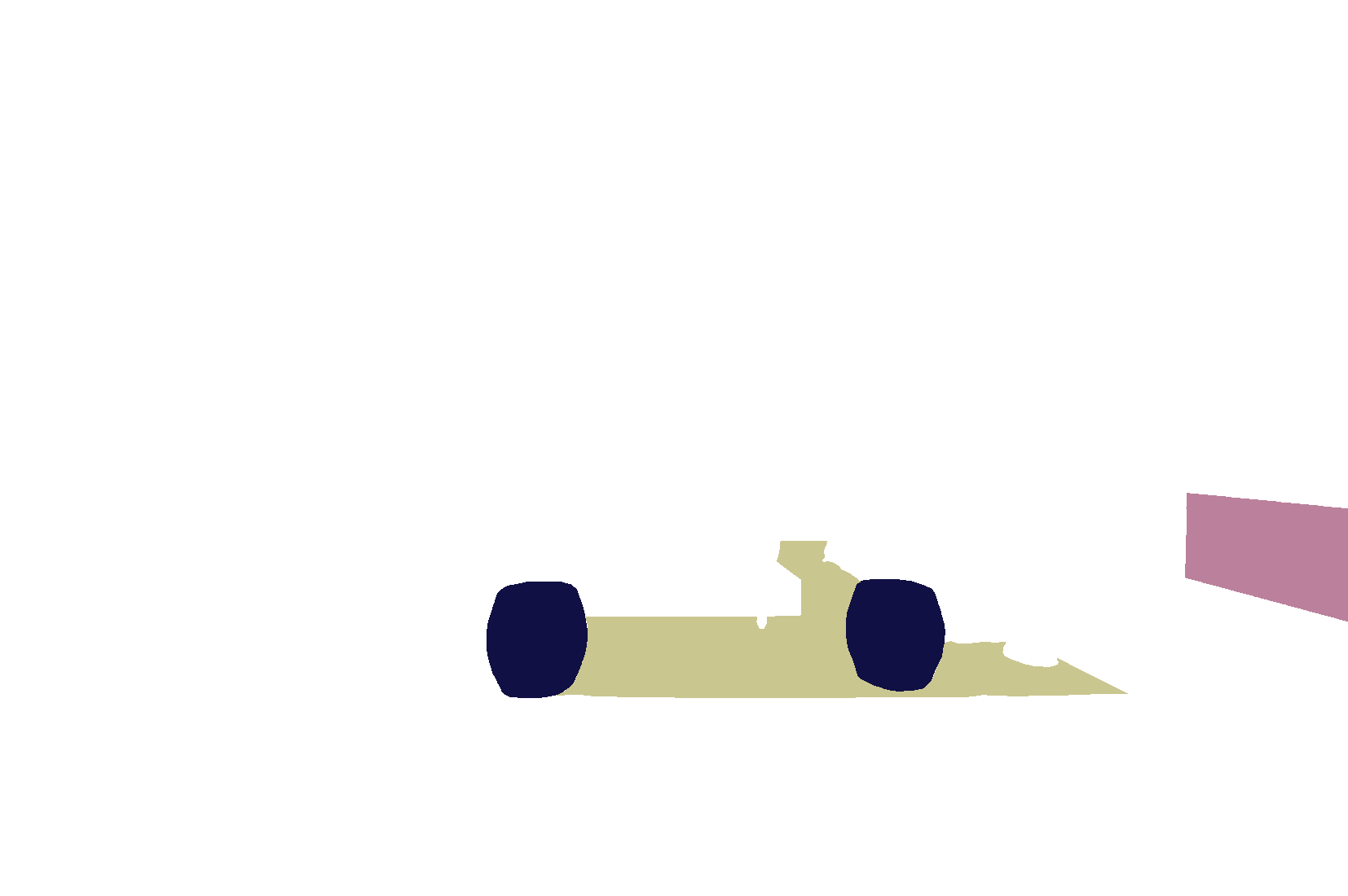}
  }
  \subfigure{%
    \includegraphics[width=.15\columnwidth]{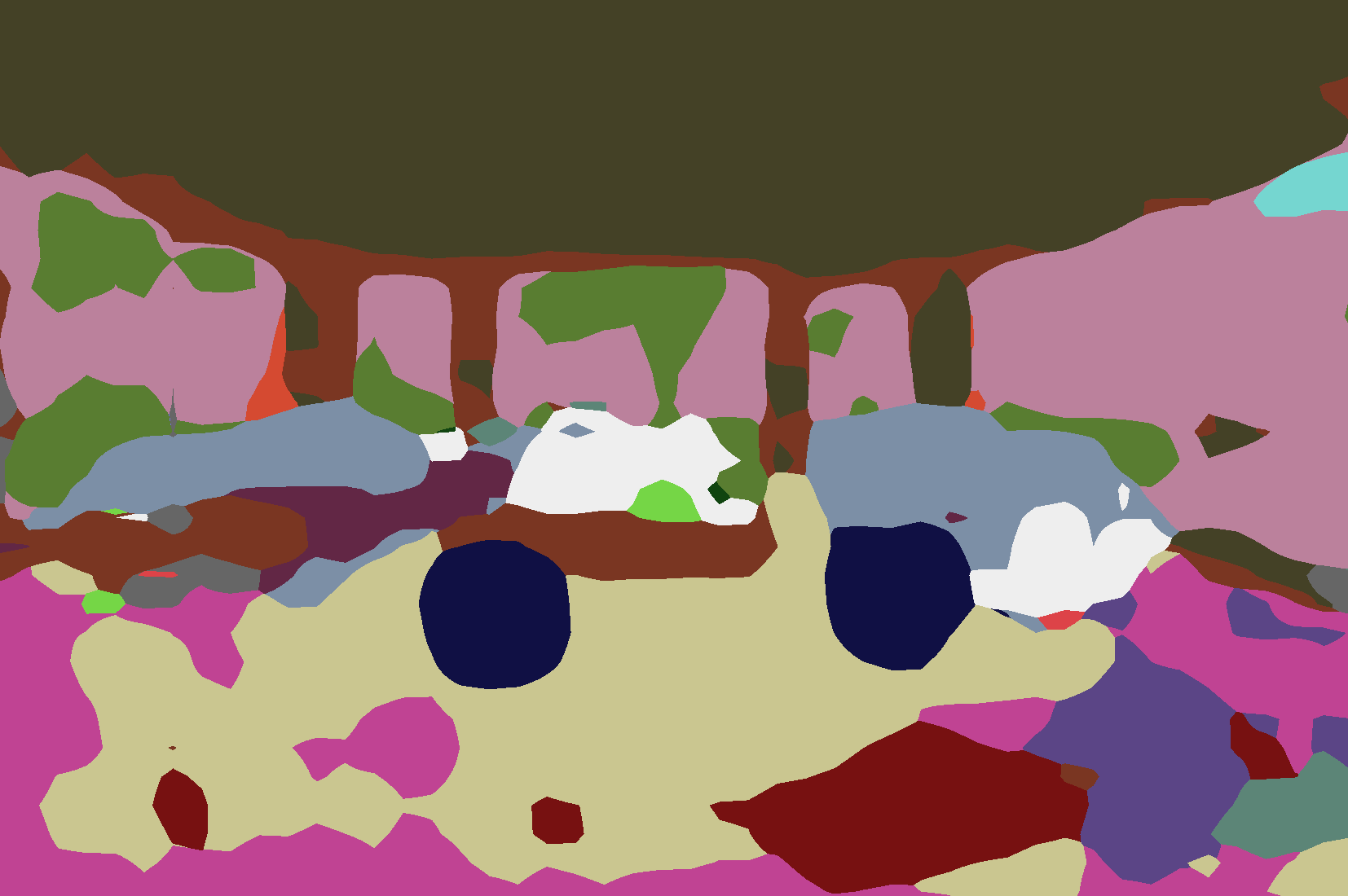}
  }
  \subfigure{%
    \includegraphics[width=.15\columnwidth]{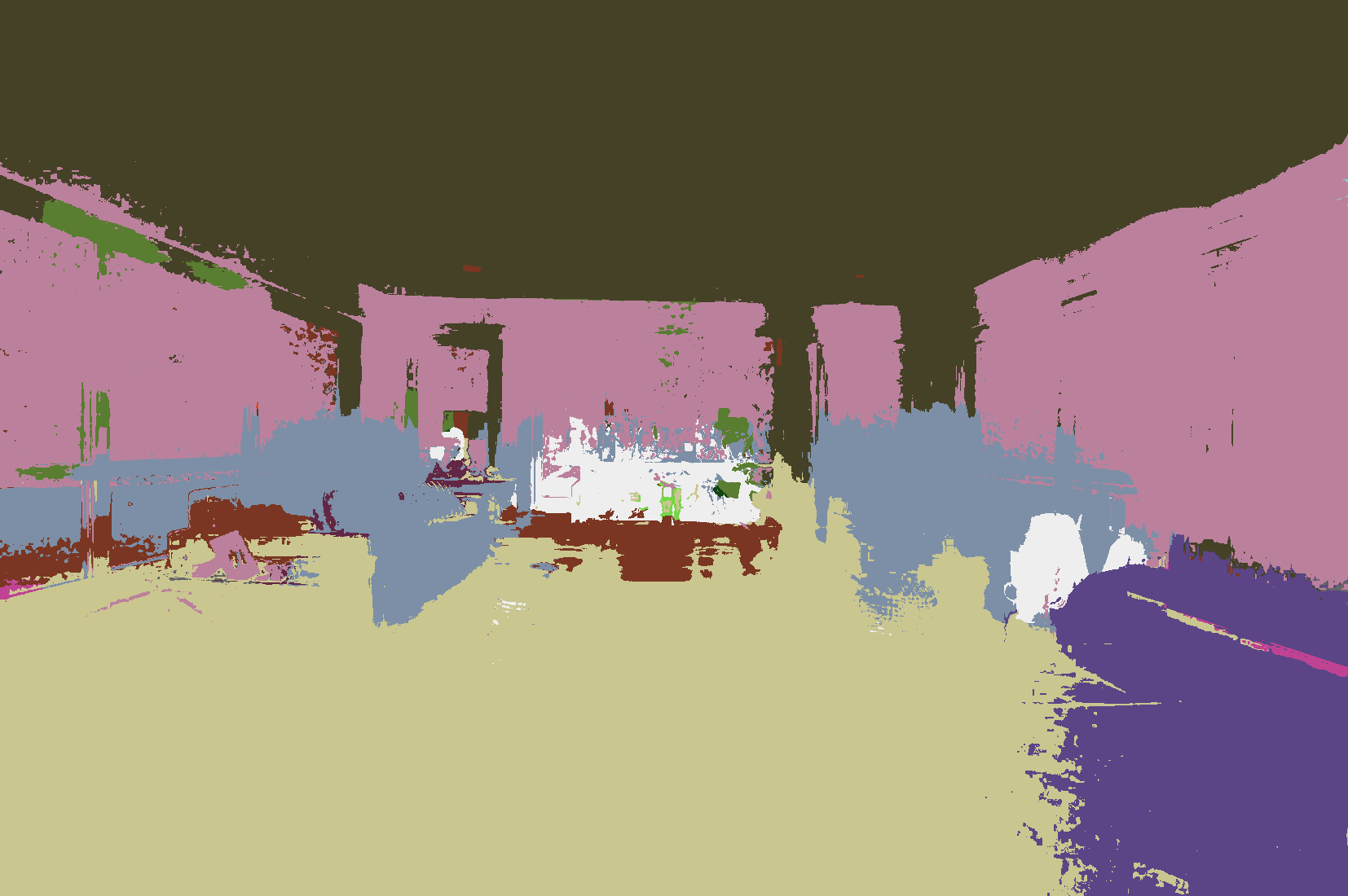}
  }
  \subfigure{%
    \includegraphics[width=.15\columnwidth]{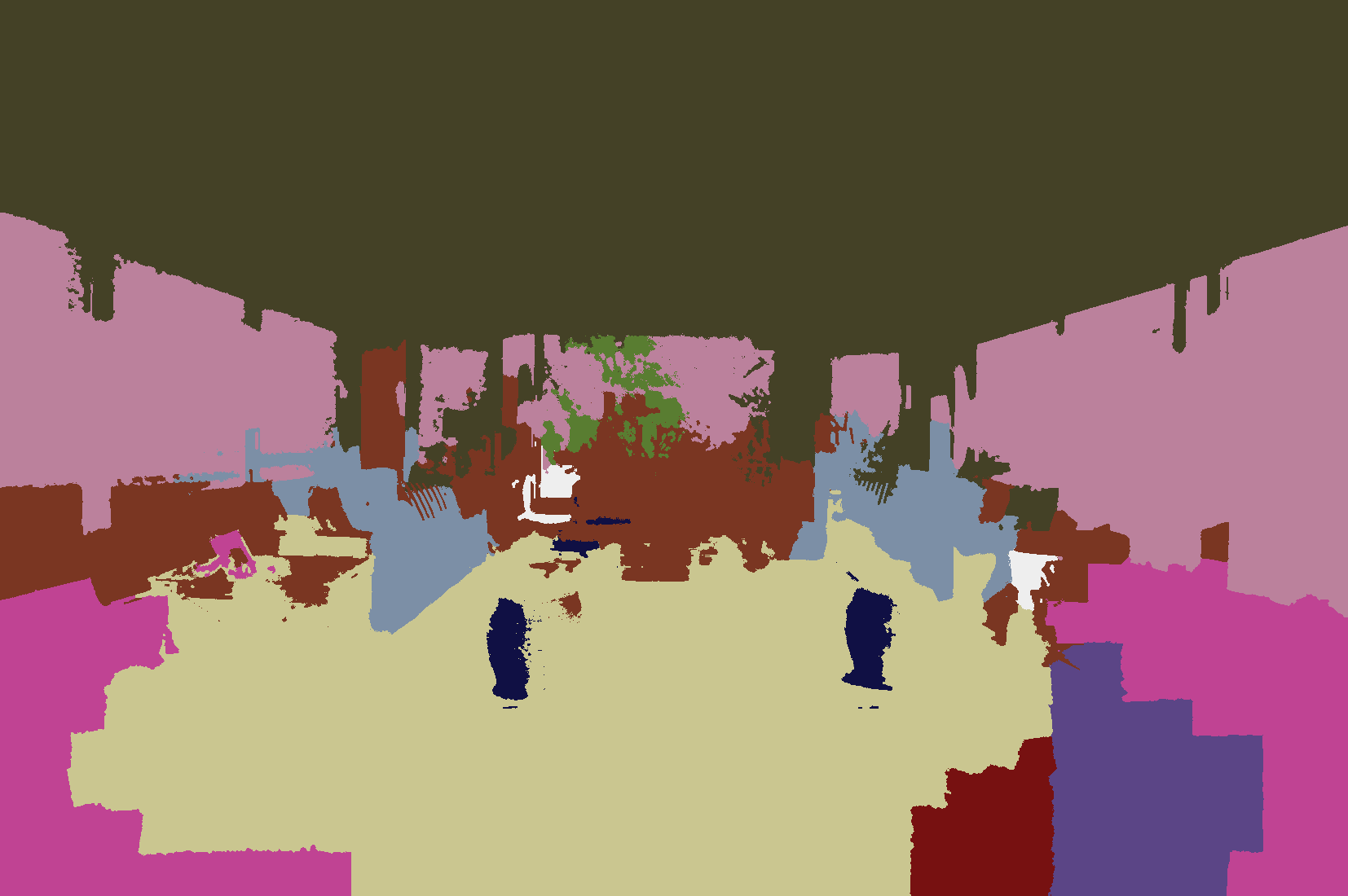}
  }\\[-2ex]
  \setcounter{subfigure}{0}
  \subfigure[\tiny Input]{%
    \includegraphics[width=.15\columnwidth]{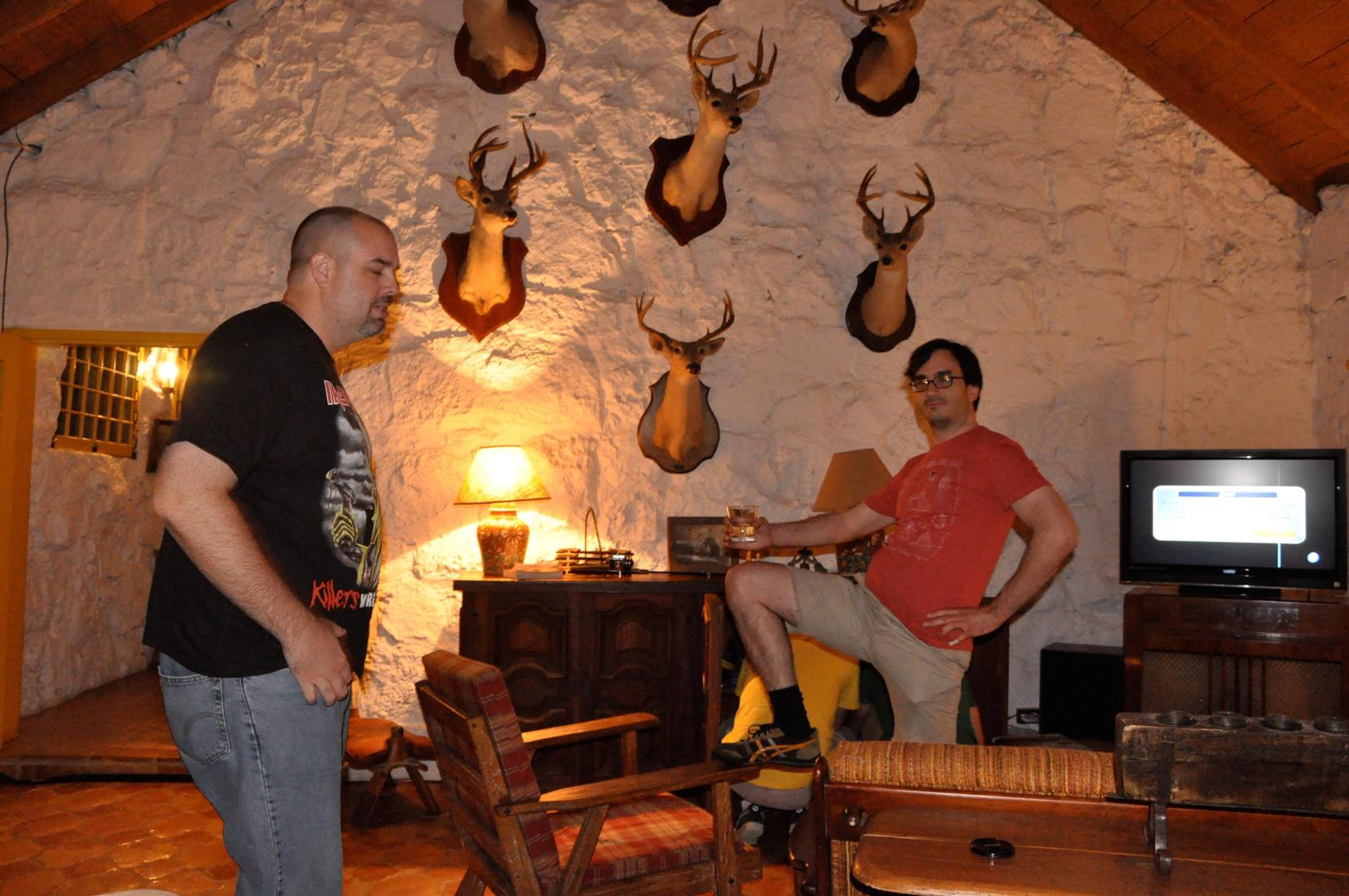}
  }
  \subfigure[\tiny Superpixels]{%
    \includegraphics[width=.15\columnwidth]{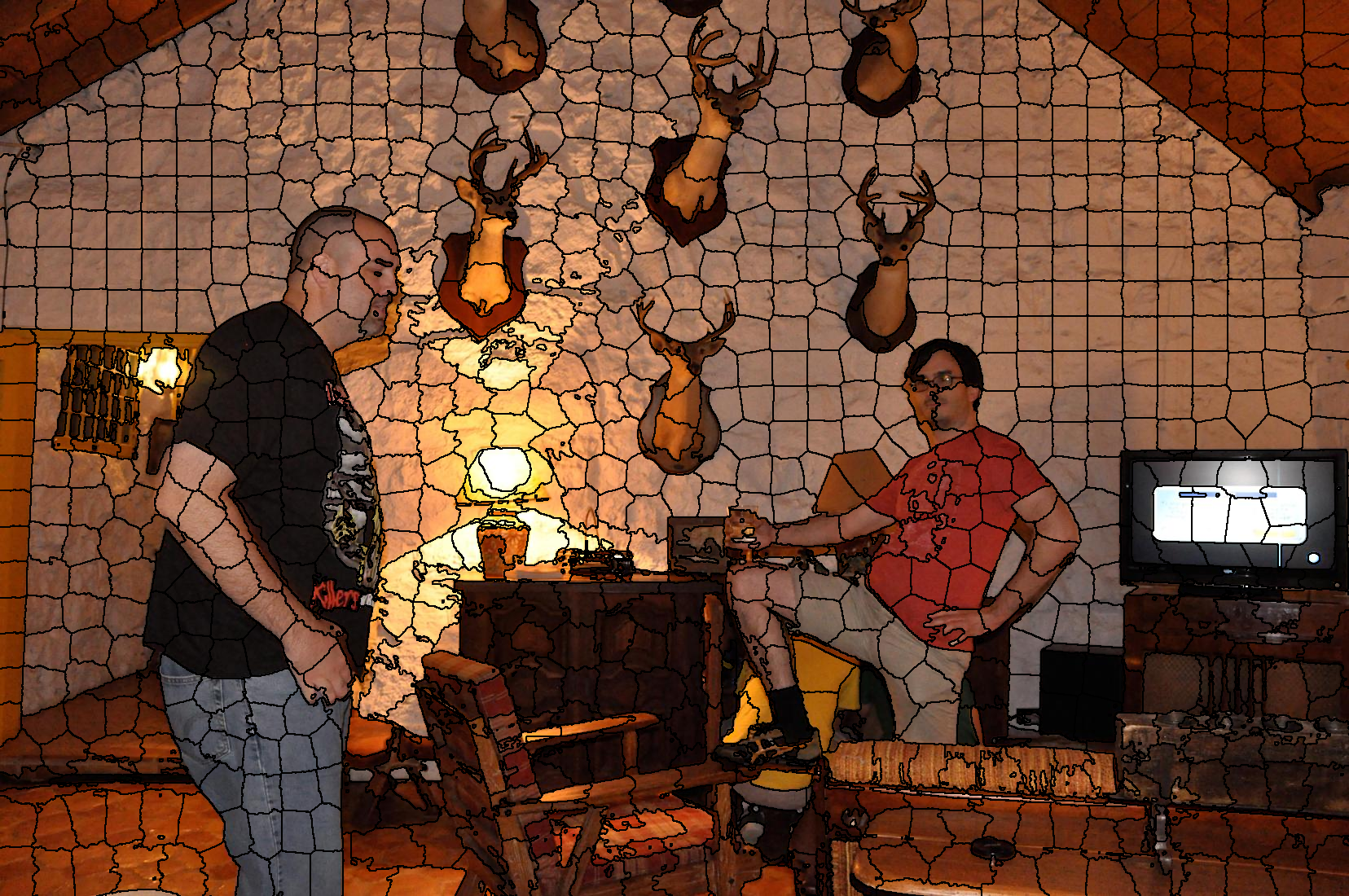}
  }
  \subfigure[\tiny GT]{%
    \includegraphics[width=.15\columnwidth]{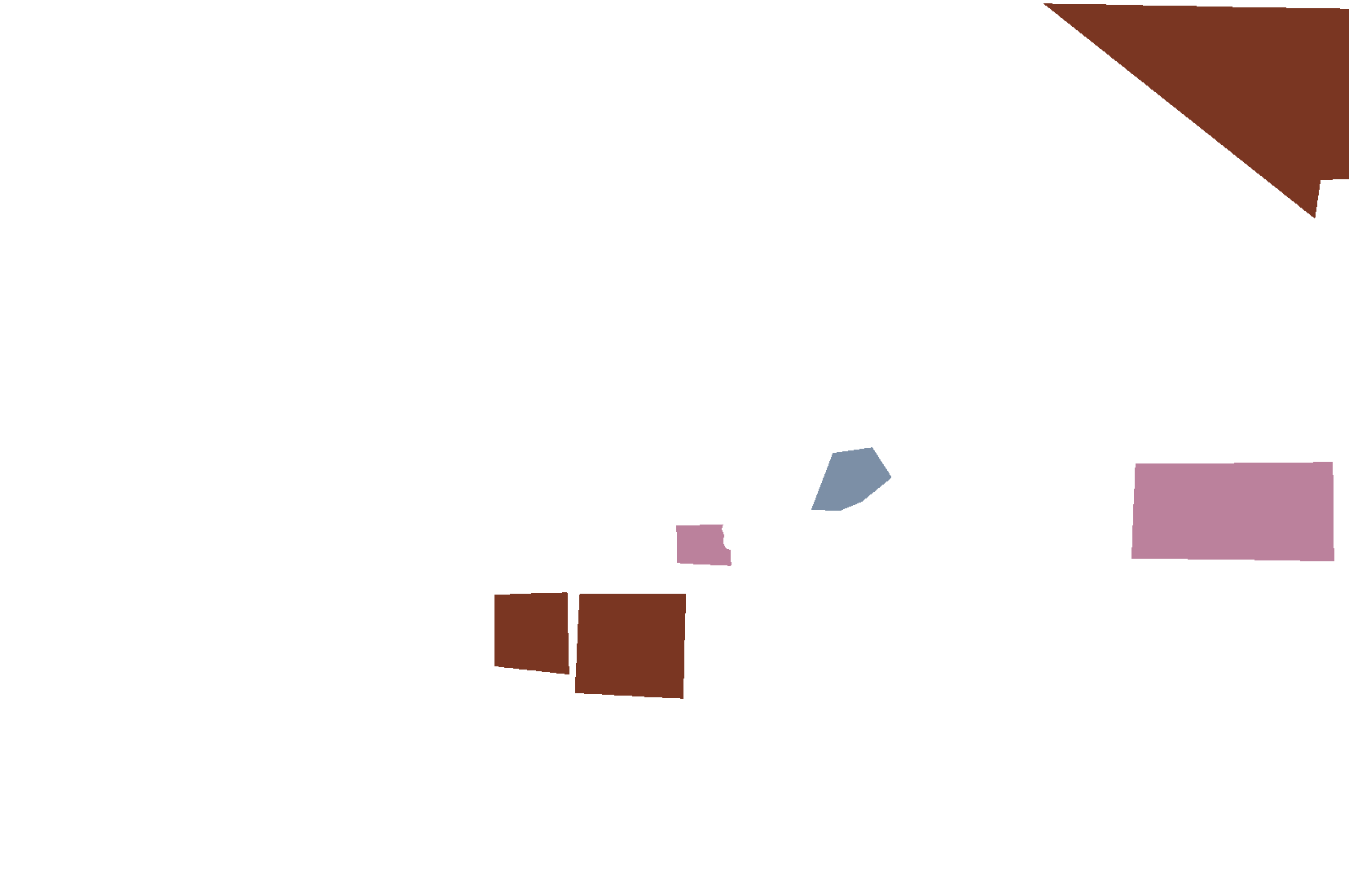}
  }
  \subfigure[\tiny AlexNet]{%
    \includegraphics[width=.15\columnwidth]{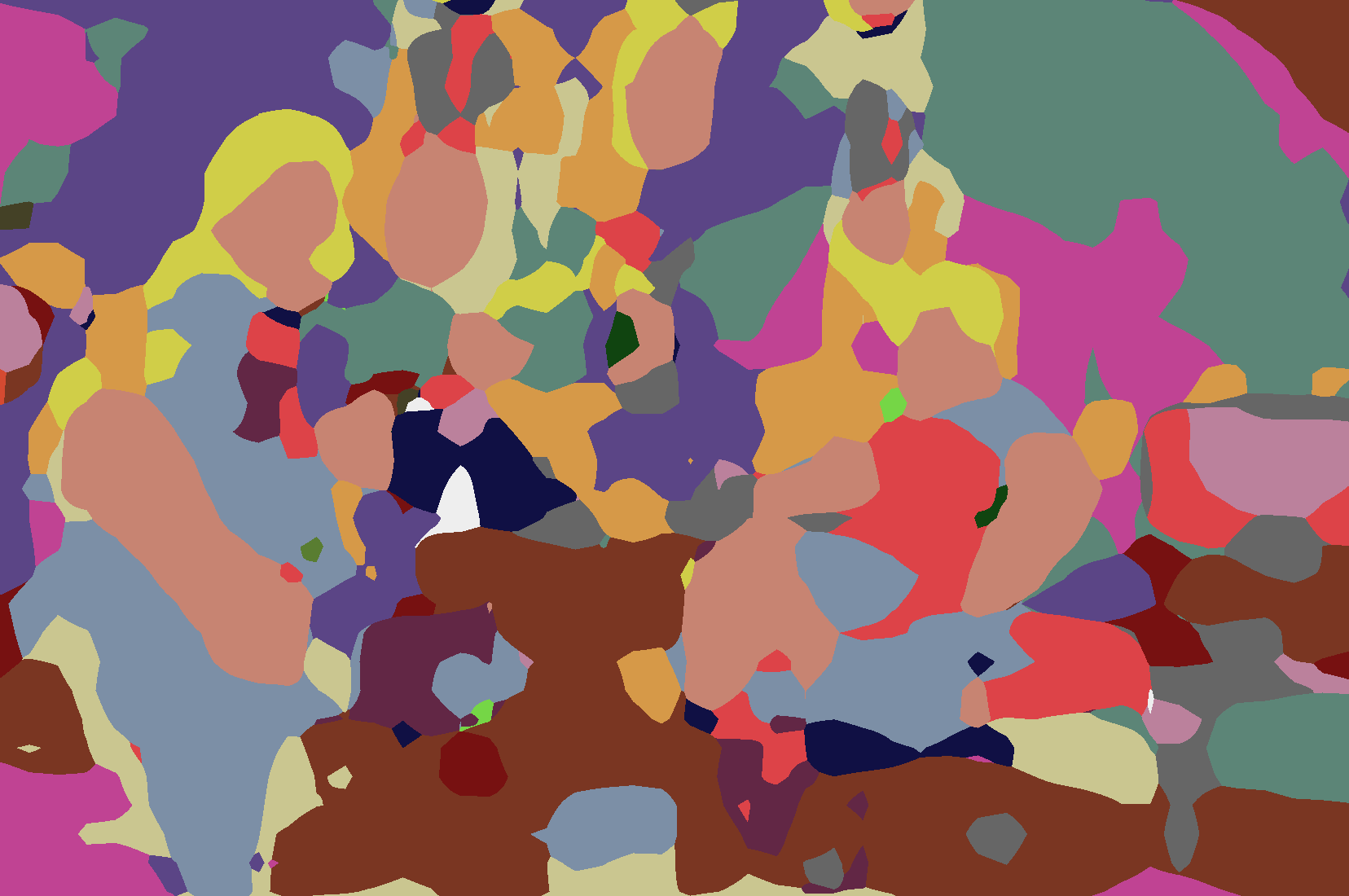}
  }
  \subfigure[\tiny +DenseCRF]{%
    \includegraphics[width=.15\columnwidth]{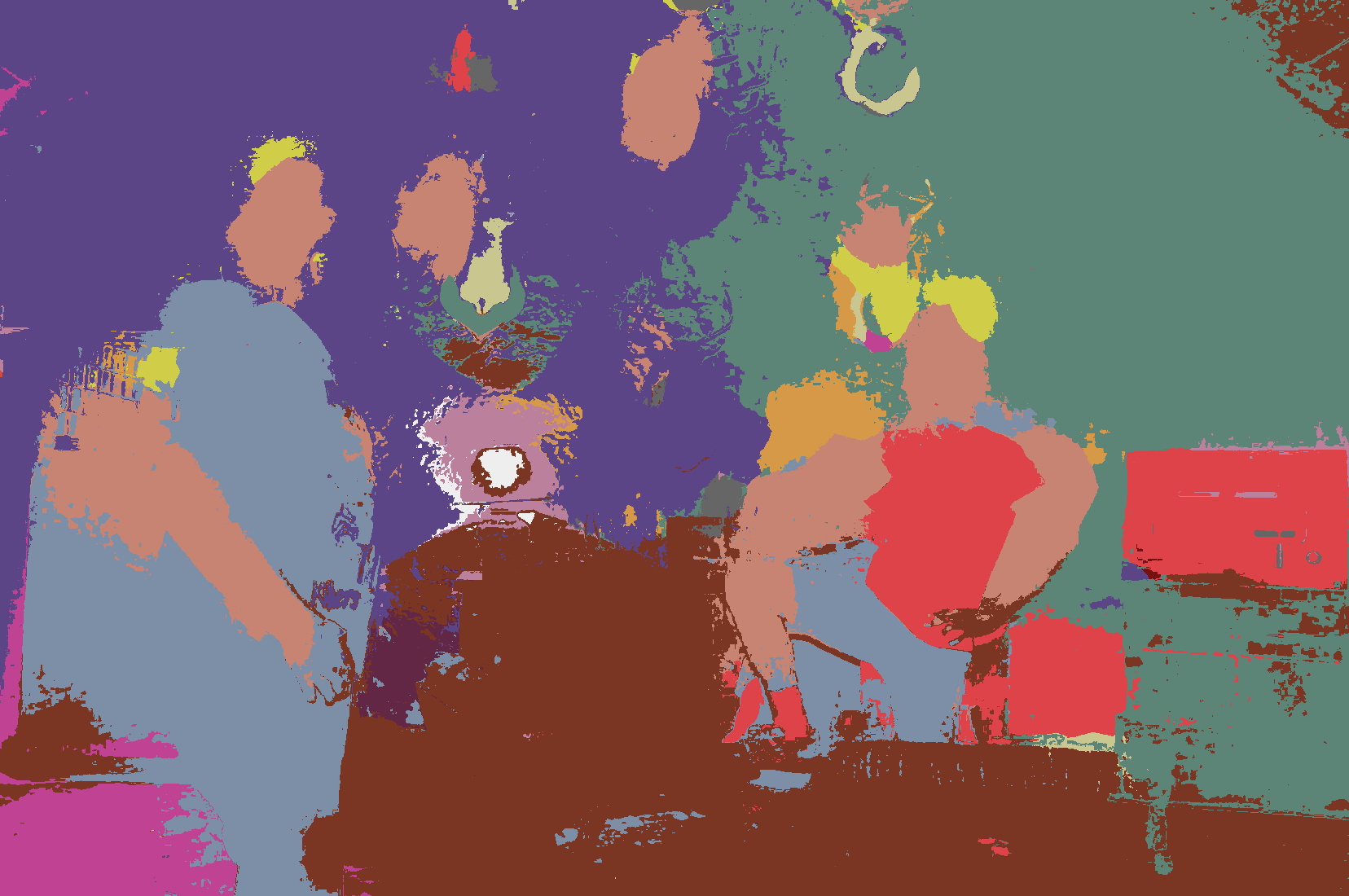}
  }
  \subfigure[\tiny Using BI]{%
    \includegraphics[width=.15\columnwidth]{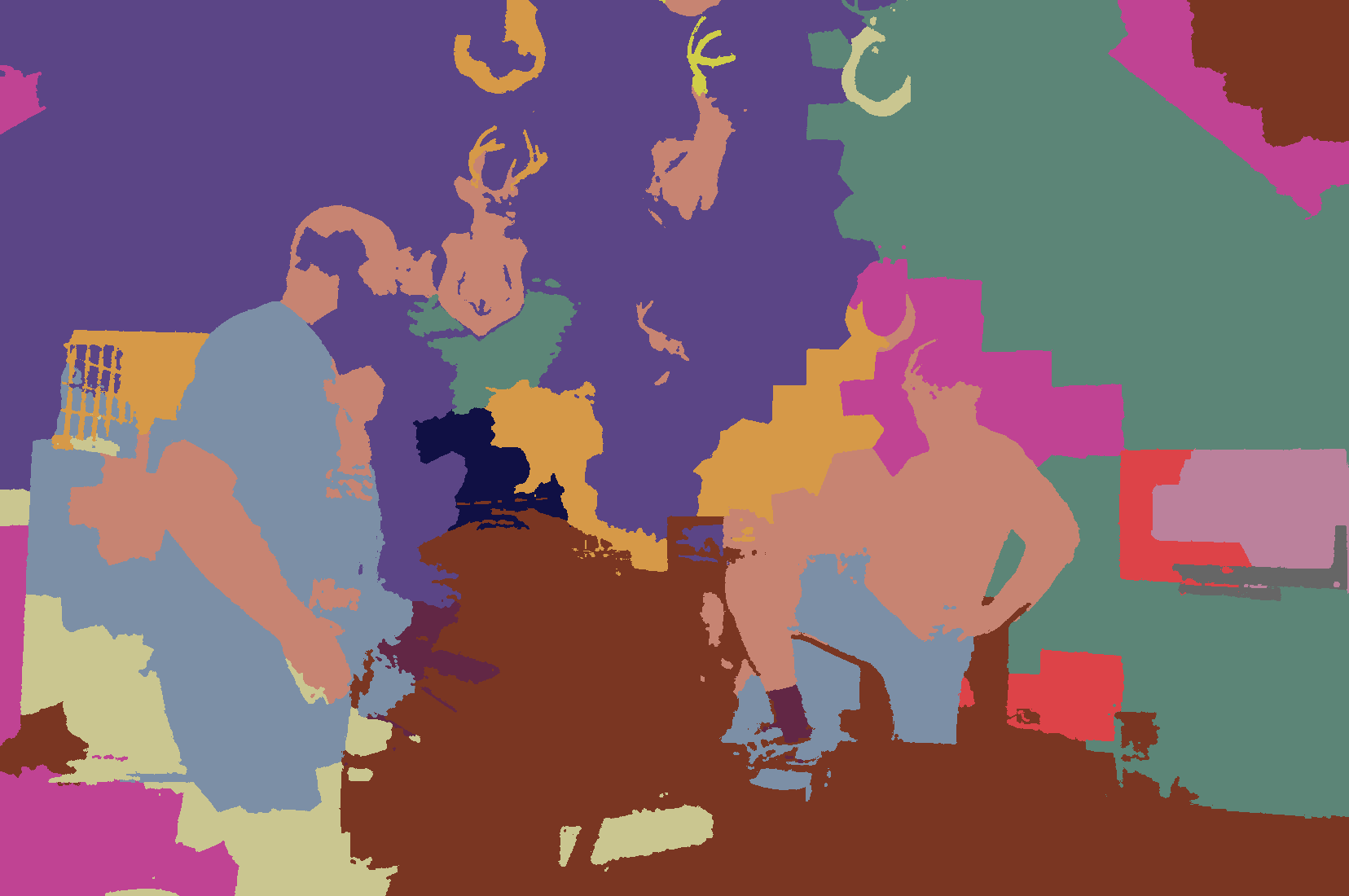}
  }
  \mycaption{Material Segmentation}{Example results of material segmentation.
  (d) depicts the AlexNet CNN result, (e) CNN + 10 steps of mean-field inference,
  (f) results obtained with bilateral inception (BI) modules (\bi{7}{2}+\bi{8}{6}) between
  \fc~layers.}
\label{fig:material_visuals}
\end{figure*}

\begin{wraptable}[16]{r}{0pt}
  \scriptsize
  \centering
  \begin{tabular}{p{1.8cm}>{\centering\arraybackslash}p{1.3cm}>{\centering\arraybackslash}p{1.3cm}>{\centering\arraybackslash}p{1.0cm}}
    \toprule
    \textbf{Model} & IoU (Half-res.) & IoU (Full-res.) & Runtime\\
    \midrule
    \scriptsize
    \deeplablargefov~CNN & 62.2 & 65.7 & 0.3s \\
    \midrule
    \bi{6}{2} & 62.7 & 66.5 & 5.7 \\
    \bi{6}{2}-\bi{7}{6} & \textbf{63.1} & \textbf{66.9} & 6.1 \\
    \midrule
    \deeplablargefovcrf & 63.0 & 66.6 & 6.9 \\
    \bottomrule
  \end{tabular}
  \mycaption{Street Scene Segmentation using \deeplablargefov~model}
  {IoU scores and runtimes (in sec) of different models on Cityscapes
  segmentation dataset~\cite{Cordts2015Cvprw}, for both half-resolution
  and full-resolution images. Runtime computations also include superpixel
  computation time (5.2s).}
  \label{tab:cityscaperesults}
\end{wraptable}

\subsection{Street Scene Segmentation}

We further evaluate the use of BI modules on the Cityscapes dataset~\cite{Cordts2015Cvprw}.
Cityscapes contains 20K high-resolution ($1024\times2048$) images of street scenes with
coarse pixel annotations and another 5K images with fine annotations, all annotations are from 19 semantic classes.
The 5K images are divided into 2975 train, 500 validation and remaining test images.
Since there are no publicly available pre-trained models for this dataset yet,
we trained a \deeplablargefov~model.
We trained the base \deeplablargefov~model with half resolution images ($512\times1024$) so that the model fits into GPU memory. The result is then interpolated to full-resolution using bilinear interpolation.

\definecolor{city_1}{RGB}{128, 64, 128}
\definecolor{city_2}{RGB}{244, 35, 232}
\definecolor{city_3}{RGB}{70, 70, 70}
\definecolor{city_4}{RGB}{102, 102, 156}
\definecolor{city_5}{RGB}{190, 153, 153}
\definecolor{city_6}{RGB}{153, 153, 153}
\definecolor{city_7}{RGB}{250, 170, 30}
\definecolor{city_8}{RGB}{220, 220, 0}
\definecolor{city_9}{RGB}{107, 142, 35}
\definecolor{city_10}{RGB}{152, 251, 152}
\definecolor{city_11}{RGB}{70, 130, 180}
\definecolor{city_12}{RGB}{220, 20, 60}
\definecolor{city_13}{RGB}{255, 0, 0}
\definecolor{city_14}{RGB}{0, 0, 142}
\definecolor{city_15}{RGB}{0, 0, 70}
\definecolor{city_16}{RGB}{0, 60, 100}
\definecolor{city_17}{RGB}{0, 80, 100}
\definecolor{city_18}{RGB}{0, 0, 230}
\definecolor{city_19}{RGB}{119, 11, 32}
\begin{figure*}[t]
  \tiny 
  \centering
  \subfigure{%
    \includegraphics[width=.18\columnwidth]{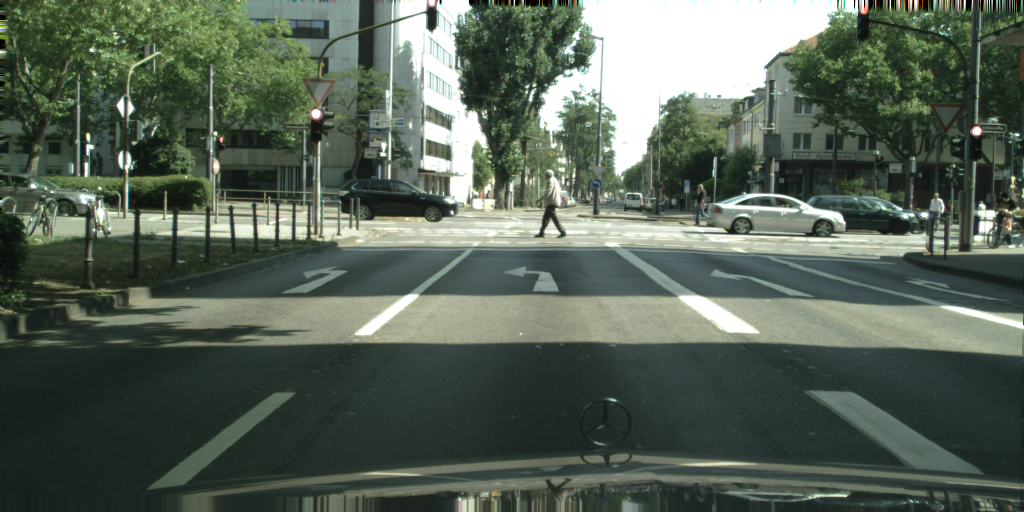}
  }
  \subfigure{%
    \includegraphics[width=.18\columnwidth]{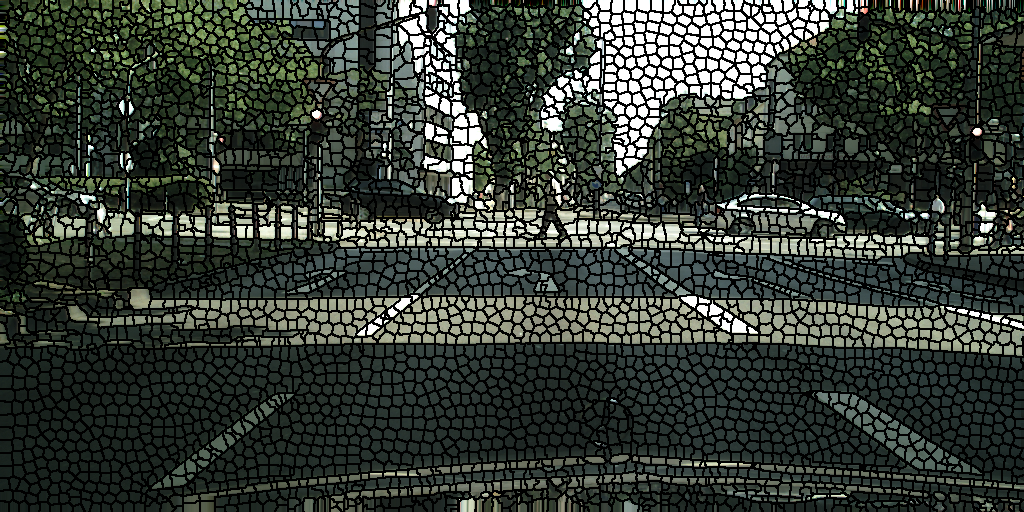}
  }
  \subfigure{%
    \includegraphics[width=.18\columnwidth]{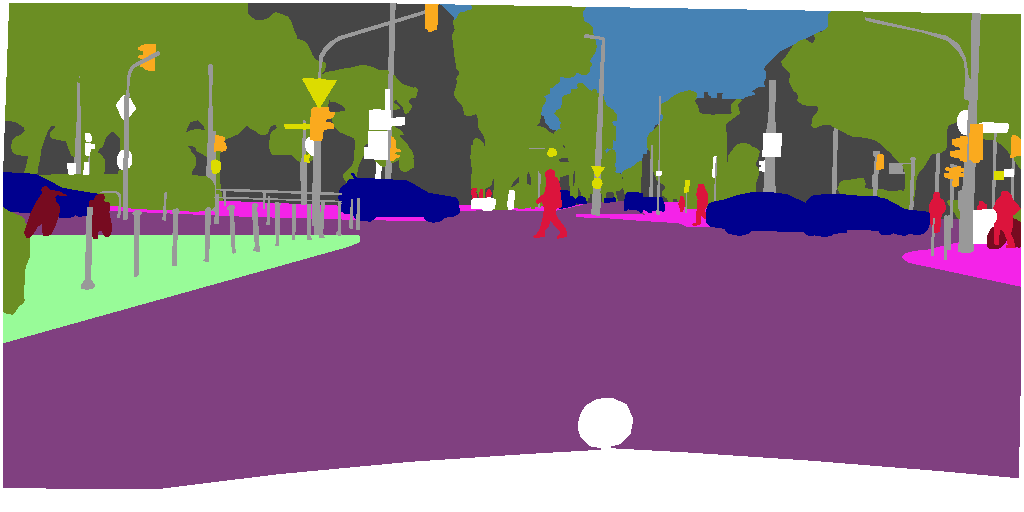}
  }
  \subfigure{%
    \includegraphics[width=.18\columnwidth]{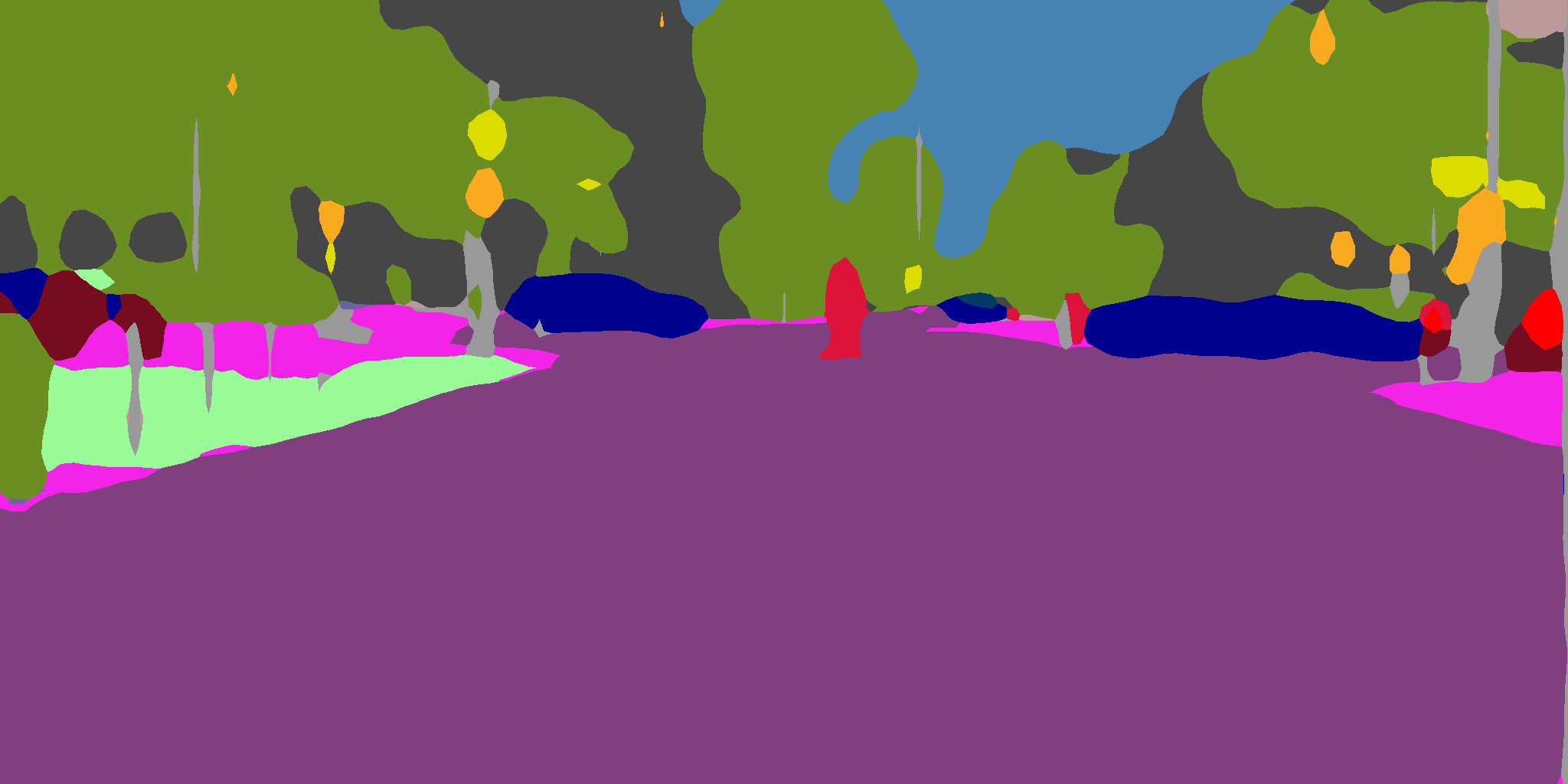}
  }
  \subfigure{%
    \includegraphics[width=.18\columnwidth]{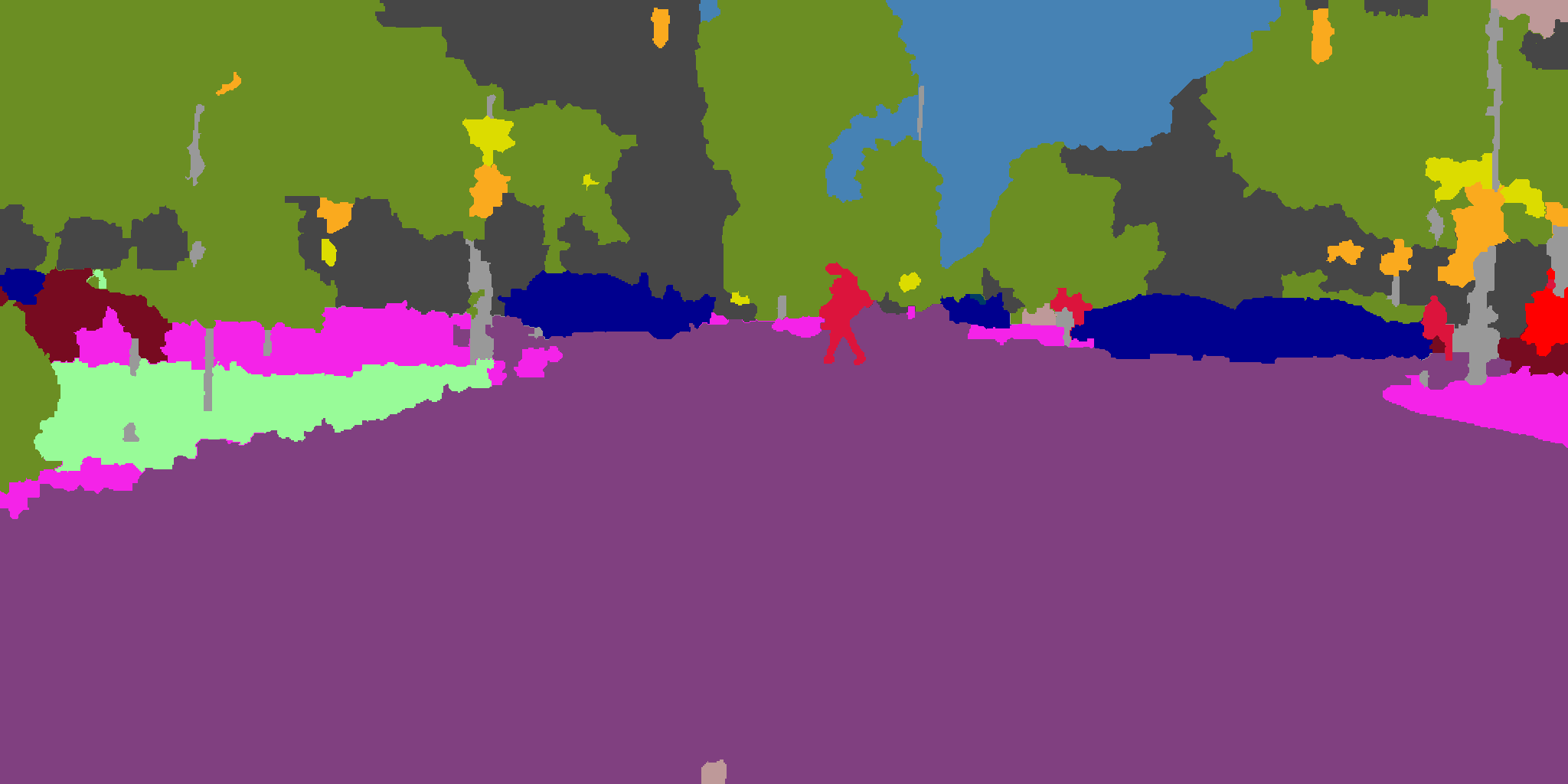}
  }\\[-2ex]
  \setcounter{subfigure}{0}
  \subfigure[\scriptsize Input]{%
    \includegraphics[width=.18\columnwidth]{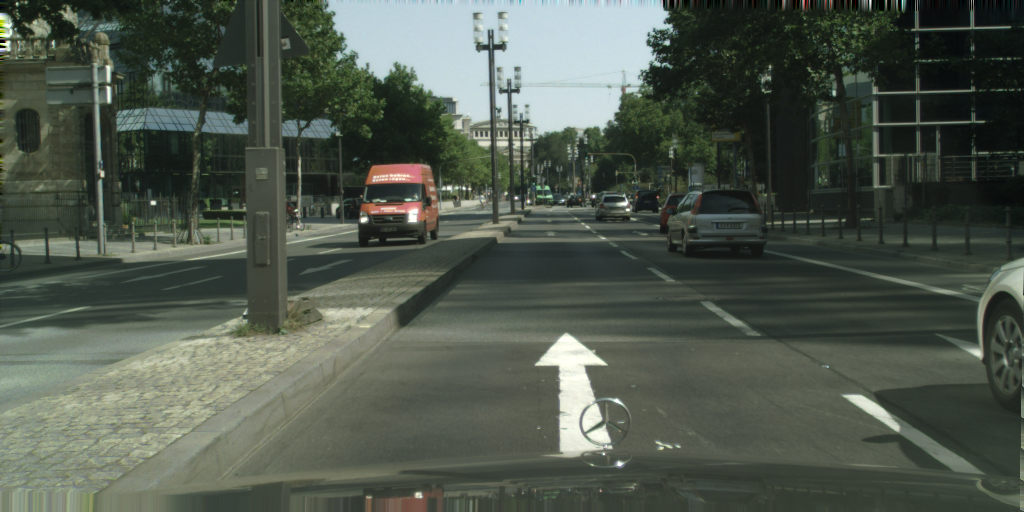}
  }
  \subfigure[\scriptsize Superpixels]{%
    \includegraphics[width=.18\columnwidth]{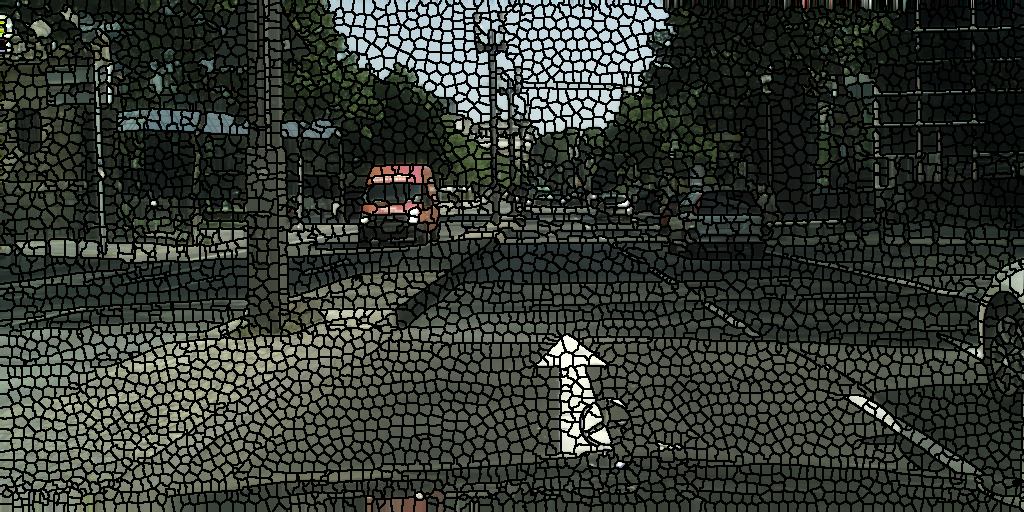}
  }
  \subfigure[\scriptsize GT]{%
    \includegraphics[width=.18\columnwidth]{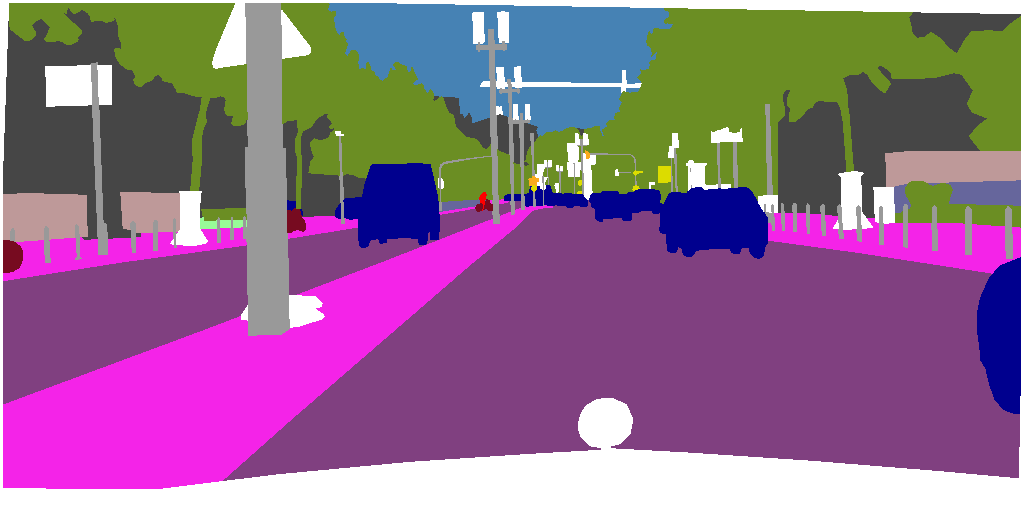}
  }
  \subfigure[\scriptsize Deeplab]{%
    \includegraphics[width=.18\columnwidth]{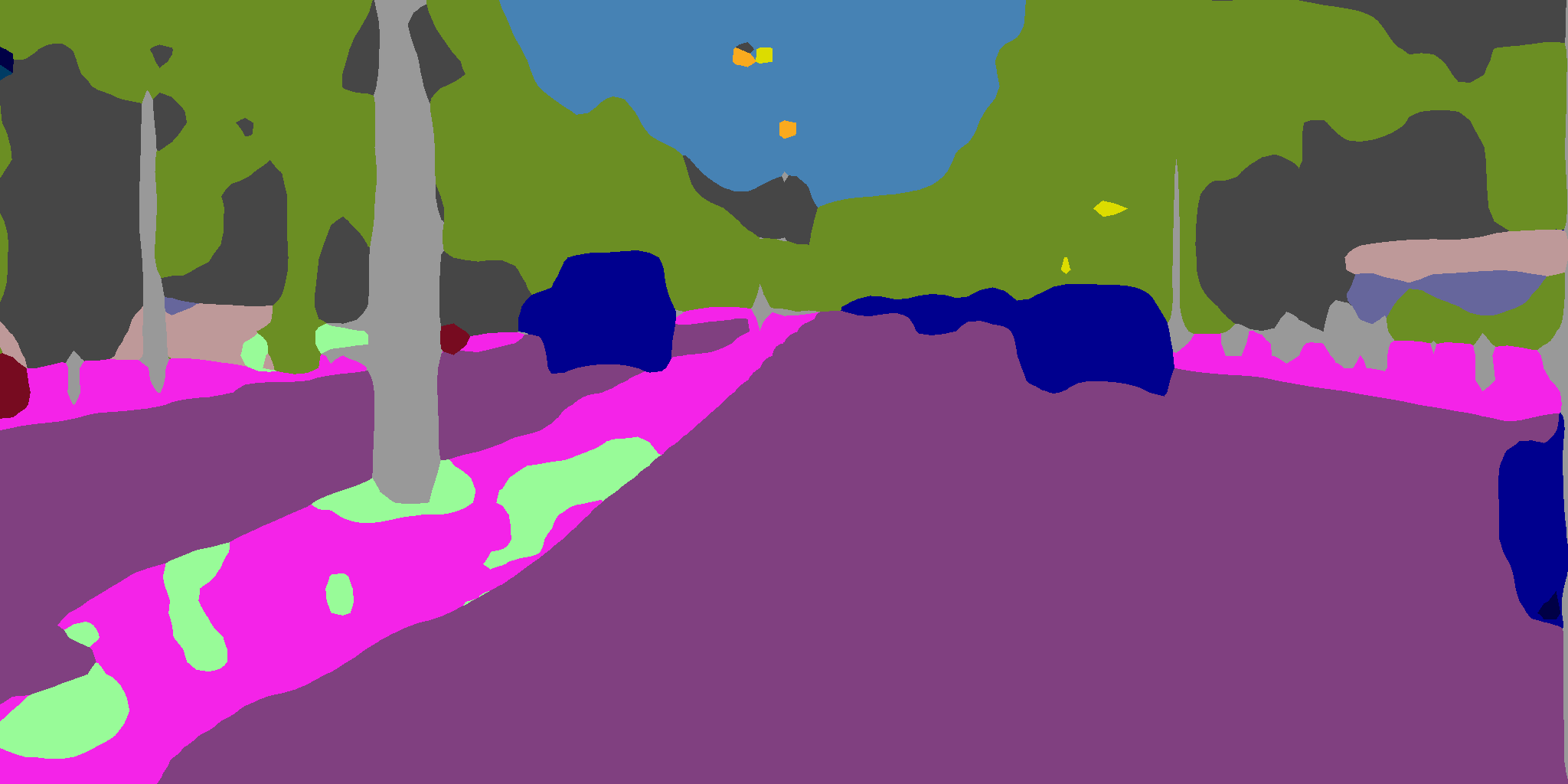}
  }
  \subfigure[\scriptsize Using BI]{%
    \includegraphics[width=.18\columnwidth]{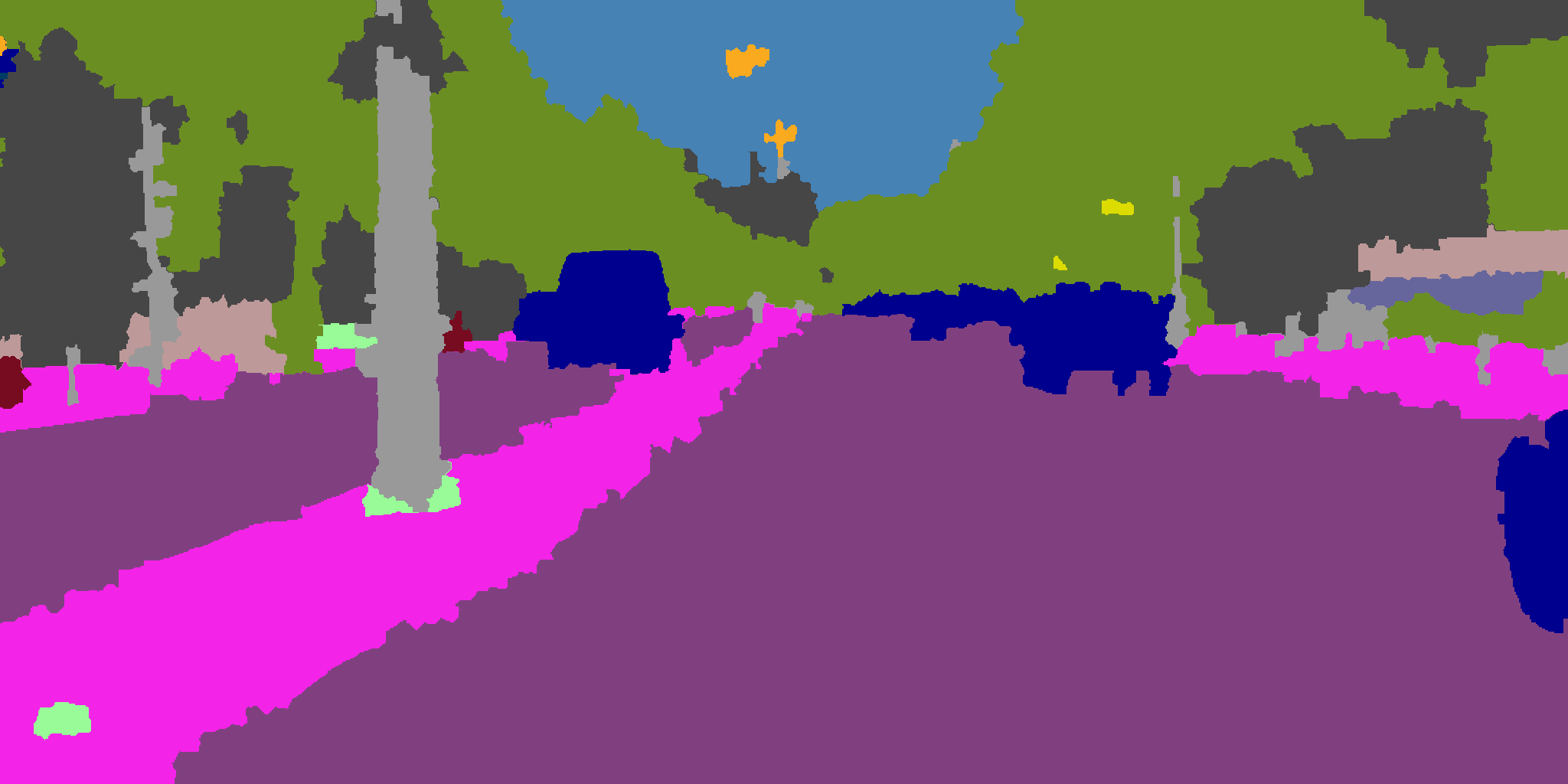}
  }
  \mycaption{Street Scene Segmentation}{Example results of street scene segmentation.
  (d) depicts the DeepLab results, (e) result obtained by adding bilateral inception (BI) modules (\bi{6}{2}+\bi{7}{6}) between \fc~layers. More in supplementary.}
\label{fig:street_visuals}
\end{figure*}

We experimented with two layouts: only a single \bi{6}{2}
and one with two inception \bi{6}{2}-\bi{7}{6} modules. We notice that the
SLIC superpixels~\cite{achanta2012slic} give higher quantization error than on VOC
and thus used 6000 superpixels using~\cite{DollarICCV13edges} for our experiments.
Quantitative results on the validation set are
shown in Tab.~\ref{tab:cityscaperesults}. In contrast to the findings on the previous datasets, we only
observe modest improvements with both DenseCRF and our inception modules in comparison
to the base model. Similar to the previous experiments, the inception modules achieve
better performance than DenseCRF while being faster. The majority of the computation
time in our approach is due to the extraction of superpixels
($5.2s$) using a CPU implementation. Some visual results with \bi{6}{2}-\bi{7}{6} model are shown in Fig.~\ref{fig:street_visuals} with more in supplementary.

\section{Conclusion}
The DenseCRF~\cite{krahenbuhl2012efficient} with mean field inference has been used in many CNN segmentation approaches.
Its main ingredient and reason for the improved performance is the use of a bilateral filter applied to the beliefs over labels. We have introduced a CNN approach that uses this key component in a novel way: filtering intermediate representations of higher levels in CNNs while jointly learning the task-specific feature spaces.
This propagates information between earlier and more detailed intermediate representations of the classes instead of beliefs over labels.
Further we show that image adaptive layouts in the higher levels of CNNs can be used to an advantage in the same spirit as CRF graphs have been constructed using superpixels in previous works on semantic segmentation.
The computations in the $1\times1$ convolution layers scales in the number of superpixels which may be an advantage. Further we have shown that the same representation can be used to interpolate the coarser representations to the full image.

The use of image-adaptive convolutions in between the FC layers retains the appealing effect of producing segmentation masks with sharp edges.
This is not a property of the superpixels, using them to represent information in FC layers and their use to interpolate to the full resolution are orthogonal.
Different interpolation steps can be used to propagate the label information to the entire image, including bilinear interpolation, up-convolutions and DenseCRFs.
We plan to investigate the effect of different sampling strategies to represent information in the higher layers of CNNs and apply similar image-adaptive ideas to videos.

We believe that the Bilateral Inception models are an interesting step that aims to directly include the model structure of CRF factors into the forward architecture of CNNs.
The BI modules are easy to implement and are applicable to CNNs that perform structured output prediction.

\small\subsubsection{Acknowledgements}
We thank the reviewers for their valuable feedback. Raghudeep Gadde is supported by CSTB and ANR-13-CORD-0003.

\small
\bibliographystyle{splncs}
\bibliography{references}

\end{document}